\pdfoutput=1
\documentclass[11pt]{article}

\usepackage[table]{xcolor}
\usepackage[final]{acl}
\definecolor{RQOneBg}{HTML}{E8F1FA}    % soft blue
\definecolor{RQOneFg}{HTML}{1F4E79}

\definecolor{RQTwoBg}{HTML}{EAF6EA}    % soft green
\definecolor{light_green}{HTML}{ddf0e5}
\definecolor{light_red}{HTML}{f7dfdf}
\definecolor{RQTwoFg}{HTML}{2E6B3A}

\definecolor{RQThreeBg}{HTML}{FFF4E5}  % soft orange
\definecolor{RQThreeFg}{HTML}{8A5A00}

\definecolor{RQFourBg}{HTML}{F3E5F5}   % soft purple
\definecolor{RQFourFg}{HTML}{4A235A}

\definecolor{DarkGreen}{HTML}{00b153}
\usepackage{times}
\usepackage{latexsym}
\usepackage{amsmath}
\usepackage{todonotes}

\usepackage{amssymb}

\usepackage[most]{tcolorbox}
\usepackage[T1]{fontenc}
\usepackage[utf8]{inputenc}

\usepackage{microtype}

\usepackage{inconsolata}

\usepackage{graphicx}

\usepackage{contour}

\usepackage{booktabs}
\usepackage{xcolor}
\usepackage{colortbl}
\usepackage{pgf}

\definecolor{DeltaNeg}{HTML}{D96C68}   % degradation
\definecolor{DeltaPos}{HTML}{6FBF95}   % improvement

\newcommand{\DeltaMax}{56}    % largest |value| in the table; sets the scale
\newcommand{\DeltaCap}{48}    % max saturation (raise to ~58 for more contrast)
\newcommand{\DeltaFlip}{200}  % |value| above which text turns white; 200 = never

\newlength{\signgap}
\AtBeginDocument{\setlength{\signgap}{\widthof{+}-\widthof{-}}}
\newcommand{\signpad}[1]{\ifdim #1pt<0pt\hspace*{\signgap}\fi}

\newcommand{\shadecell}[1]{%
  \pgfmathsetmacro{\magv}{min(\DeltaCap,abs(#1)/\DeltaMax*100)}%
  \ifdim #1pt<0pt
    \edef\nextcell{\noexpand\cellcolor{DeltaNeg!\magv!white}}%
  \else
    \edef\nextcell{\noexpand\cellcolor{DeltaPos!\magv!white}}%
  \fi
  \nextcell
}
\newcommand{\txtcol}[1]{%
  \pgfmathsetmacro{\magt}{abs(#1)}%
  \ifdim\magt pt>\DeltaFlip pt \color{white}\fi
}
\newcommand{\D}[1]{\shadecell{#1}\txtcol{#1}\signpad{#1}#1\phantom{*}}
\newcommand{\Ds}[1]{\shadecell{#1}\txtcol{#1}\signpad{#1}#1*}

\newcommand{\hide}[1]{}

\newcommand{\TAMARBUTA}{{$\hbar$}}

\newcommand{\SHIN}{{\v{s}}}
\usepackage{arabtex}  
\usepackage{hebtex}  
\usepackage{textgreek}
\usepackage[utf8]{inputenc} % Support UTF-8 input
\DeclareUnicodeCharacter{0127}{\hbar}

\usepackage{booktabs}
\usepackage{multirow}
\usepackage{utf8}
\usepackage{epstopdf}
\usepackage{pdfpages}

\title{When Similar Means Different: \\ Evaluating LLMs on Arabic--Hebrew Cognates}
\author{
Junhong Liang \quad Noor Abo Mokh \quad Bashar Alhafni\\
Mohamed bin Zayed University of Artificial Intelligence\\
\texttt{\{junhong.liang,noor.abomokh,bashar.alhafni\}@mbzuai.ac.ae}
}

\begin{document}
\maketitle

\setcode{utf8}
\setarab
\vocalize

%\textbf{\textsc{Hello}}
%\textsc{Hello} 

\begin{abstract}
Arabic and Hebrew, as closely related Semitic languages, share many words with similar surface forms, including true cognates, false friends, and modern loanwords. These lexical relationships create ambiguity for cross-lingual semantic interpretation, as similar forms may correspond to shared, divergent, or borrowed meanings.
To evaluate whether LLMs can make these distinctions, we introduce SemCog Bench, a curated benchmark of 1,858 Arabic--Hebrew word pairs with sentence-level annotations for cognate identification and semantic disambiguation.  We evaluate a diverse set of open-source and proprietary LLMs across multiple input representations and contextual settings.  Our results show reliance on surface-form similarity, with weaker performance on false friends and wide variation on loanwords. We further find that context and scale yield model-dependent gains, while original-script inputs generally perform best.
% We evaluate open-source and commercial LLMs across multiple input representations (raw, diacritized, Romanized, and phonetic) and  reveal a critical gap in cross-lingual reasoning. While models achieve high accuracy on true cognates, performance drops sharply on false friends and loanwords, reflecting a strong reliance on surface-form similarity. Furthermore, sentence-level context yields only modest improvements, suggesting that contextual cues alone are insufficient to overcome misleading form-based signals.
% Performance also drops substantially with phonetic representations, highlighting a dependence on script-based pretraining signals. 
% These findings reveal a fundamental limitation of current LLMs in resolving cross-lingual form-meaning conflicts and establish SemCog Bench as a rigorous benchmark for multilingual semantic reasoning.
Our findings reveal limitations in cross-lingual form-meaning reasoning and establish SemCog Bench as a benchmark for cross-lingual lexical semantics.
Our code and data are publicly available.\footnote{\url{https://github.com/mbzuai-nlp/SemCog}}

% Furthermore, providing sentence-level context fails to insulate models from form--meaning conflicts, and phonetic representations severely disrupt script-based pretraining alignments. Ultimately, our findings demonstrate that current LLMs struggle to operationalize deep, language-specific semantic representations to overcome cross-lingual interference, establishing SemCog Bench as a rigorous diagnostic tool for evaluating multilingual robustness.

\end{list} % DON'T REMOVE

% \begin{tabular}{l}
% \hspace{2em}\raisebox{-0em}{\includegraphics[height=1em]{github.png}} \href{https://github.com/mbzuai-nlp/SemCog}{\texttt{mbzuai-nlp/SemCog}} \\
% \end{tabular}

\end{abstract}

\section{Introduction}
\label{sec:intro}
% https://docs.google.com/presentation/d/1CqsZYLrezbCvSlef4QJTTc1zdJGbmwLjFf6S3leY0k8/edit?slide=id.g3ea27a1006d_0_20#slide=id.g3ea27a1006d_0_20
% https://drive.google.com/file/d/1JMaeq7p7XzPCH_gLJDA3uVSv6_rS8tsv/view?usp=drive_link
\begin{figure}[!t]
    \centering
    \includegraphics[width=\linewidth]{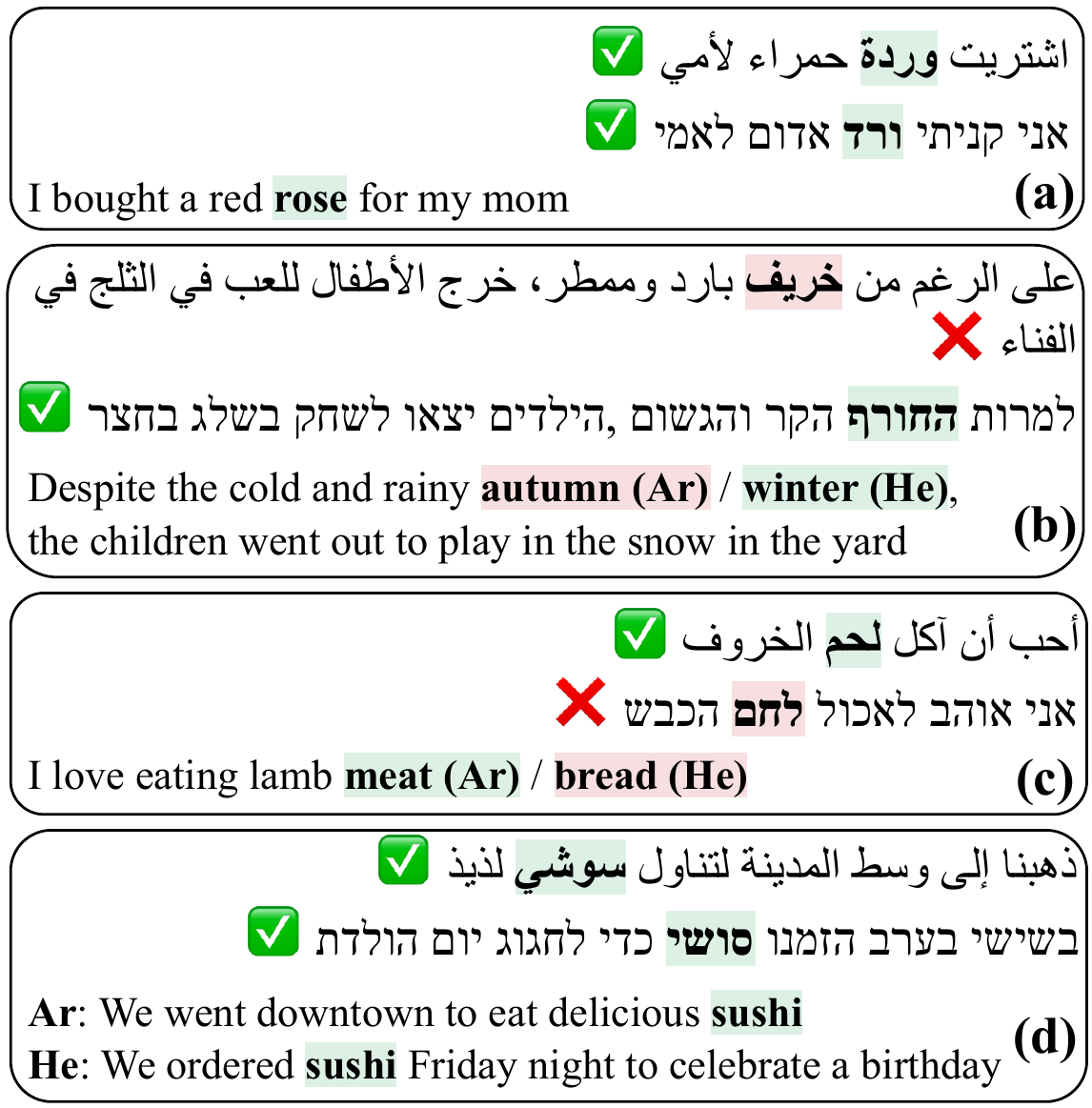}
    \caption{Examples of Arabic and Hebrew true cognates (a), false friends (b, c) and loanwords (d). Correct usages are marked in \textcolor{DeltaPos}{\textbf{green}} and incorrect usages in \textcolor{DeltaNeg}{\textbf{red}}.}
    \label{fig:example_ar_he}
\end{figure}

% Semitic languages, such as Arabic and Hebrew, share common morphological roots and exhibit substantial lexical overlap. This overlap includes a large number of cognates with similar orthographic and phonological forms.

Arabic and Hebrew, as closely related Semitic languages, exhibit substantial lexical overlap, including many words with similar surface forms. While some of these words are \textit{true cognates} that preserve semantically related meanings across the two languages, others are \textit{false friends} that exhibit similar orthographic and phonological forms but differ in meaning~\cite{cahyawijaya2025thank}. A third category consists of \textit{loanwords} borrowed from other languages, which may also appear highly similar in Arabic and Hebrew. This lexical overlap creates substantial ambiguity, since similar forms across Arabic and Hebrew may differ in both lexical relationship and meaning.
% These three lexical relationships pose a unique challenge for cross-lingual semantic understanding, as orthographic and phonological similarity can be highly misleading without deeper semantic reasoning.

% While many of these cognates preserve the same meaning across languages, others have undergone semantic divergence, resulting in \textit{false friends}---words that appear orthographically and phonologically similar but differ in meaning~\cite{cahyawijaya2025thank}. Furthermore, modern cross-cultural exchange has introduced a third category: \textit{loanwords}, borrowed from non-Semitic sources. This trifecta poses a unique challenge for cross-lingual semantic understanding, as surface-form similarity can be highly misleading without deeper semantic reasoning.

Figure~\ref{fig:example_ar_he} illustrates examples of these lexical relationships. True cognates, such as the Arabic word \setarab<وردة> (\textit{wrd{\TAMARBUTA}}\footnote{Arabic HSB transliteration~\cite{Habash:2007:arabic-transliteration}.}, `rose') and the Hebrew word \sethebrew\<ורד> (\textit{wrd}\footnote{Hebrew transliteration following \newcite{shilon-etal-2010-machine}.}, `rose'), share similar forms and meanings across the two languages (Figure~\ref{fig:example_ar_he}(a)). In contrast, false friends preserve strong surface similarity despite semantic divergence; for example, the Arabic word \setarab<خريف> (\textit{xryf}, `autumn') closely resembles the Hebrew word \sethebrew\<חורף> (\textit{xwrp}, `winter') (Figure~\ref{fig:example_ar_he}(b)).
% Similarly, \setarab<ريب> (\textit{ryb}, `doubt') and \sethebrew\<ריב> (\textit{rib}, `quarrel') are phonologically and orthographically similar yet encode entirely distinct meanings (Figure~\ref{fig:example_ar_he}(c)).
% Finally, loanwords present surface similarity driven by external borrowing rather than shared ancestry; for instance, the Japanese culinary term for `sushi' appears almost identically in both Arabic (\setarab<سوشي>, \textit{sw{\SHIN}y}) and Hebrew (\sethebrew<סושי>, \textit{sw{\SHIN}y}) (Figure~\ref{fig:example_ar_he}(d)).
Loanwords may likewise appear similar, but for a different reason: \setarab<سوشي> (\textit{sw{\SHIN}y}, `sushi') and \sethebrew<סושי> (\textit{sw{\SHIN}y}, `sushi') (Figure~\ref{fig:example_ar_he}(d)) are both modern borrowings rather than inherited Semitic cognates.
% These cases highlight that surface similarity alone is insufficient for correct interpretation; models must rely on deeper semantic understanding and etymological context to accurately disambiguate true cognates, false friends, and loanwords.
These cases demonstrate the different lexical relationships that can underlie similar forms across Arabic and Hebrew.

In this work, we introduce SemCog Bench, a curated benchmark of 1,858 Arabic--Hebrew true cognates, false friends, and loanwords with sentence-level annotations for cognate identification and semantic disambiguation. Using SemCog Bench, we evaluate a diverse set of LLMs, including open-source and proprietary multilingual models as well as Arabic- and Hebrew-centric models. We address four research questions: \colorbox{RQOneBg}{\textbf{RQ1:}} How accurately do LLMs perform Arabic--Hebrew cognate identification and semantic disambiguation? \colorbox{RQTwoBg}{\textbf{RQ2:}} How does input representation (raw, diacritized, Romanized, or phonetic) affect performance? \colorbox{RQThreeBg}{\textbf{RQ3:}} Does sentence-level context improve cognate identification? \colorbox{RQFourBg}{\textbf{RQ4:}} To what extent does model scaling improve performance on SemCog Bench? Our main contributions are as follows:
% \begin{enumerate}
% \item We introduce SemCog Bench, a curated benchmark disambiguating Arabic--Hebrew cognate pairs, comprising true cognates, false friends, and loanwords used for cognate identification and semantic disambiguation.
% \item We systematically evaluate the performance of a broad range of LLMs on cognate identification and semantic disambiguation, revealing consistent performance gaps between multilingual and language-specific models.
% \item Through controlled perturbation experiments, we demonstrate that model errors on false friend pairs are driven primarily by cross-lingual surface similarity rather than by orthographic complexity or morphological ambiguity.
% \end{enumerate}

\begin{enumerate}
\item We introduce SemCog Bench, a benchmark of Arabic--Hebrew true cognates, false friends, and loanwords for cognate identification and semantic disambiguation.
\item We evaluate a broad range of LLMs using SemCog Bench, across input representations, context settings, and model scales.
\item We show that LLMs exhibit reliance on surface-form similarity, struggling to distinguish false friends from true cognates, while loanword performance varies across models.
\item We find that context and scale provide model-dependent gains, and that original-script input generally outperforms other representations.
\end{enumerate}

% background and related work. 

\section{Background and Related Work}
\label{sec:related_work}
\subsection{Arabic--Hebrew Linguistic Background}
Arabic and Hebrew are closely related Semitic languages characterized by rich morphology and orthographic ambiguity. Both employ root-and-pattern word formation, where words are derived by combining a consonantal root, often tri-consonantal, with vocalic patterns, resulting in large and sparse vocabularies. In addition, both languages use optional diacritics to represent short vowels--\textit{tashkil} in Arabic and \textit{nikkud} in Hebrew--that are typically omitted in modern text~\citep{Habash:2010:introduction,seddah-etal-2013-overview}. As a result, many surface forms are ambiguous, posing challenges for computational modeling \citep{tsarfaty-etal-2020-spmrl}.

At the same time, the shared Semitic ancestry of Arabic and Hebrew gives rise to numerous lexical parallels, including words with similar orthographic or phonological forms. These include true cognates, false friends, and loanwords. Cognates are words in different languages that share a common historical origin \citep{lijewska2020cognate,abu2025semantic}. When cognates preserve both form and meaning across the two languages, they constitute true cognates (Figure~\ref{fig:example_ar_he}(a)); when their forms are similar but their meanings diverge, they may result in false friends \cite{dominguez-false-friends}. Hebrew \sethebrew\<לחם> (\textit{lxm}, `bread') and Arabic \setarab\<لحم>\verb| |(\textit{lHm}, `meat'), for example, both derive from the Semitic root \textit{l-H-m} yet denote different things (Figure~\ref{fig:example_ar_he}(c)). Loanwords (Figure~\ref{fig:example_ar_he}(d)), instead, owe their similarity to parallel borrowing rather than common descent. Although the term generally refers to words borrowed from one language into another~\citep{mahajna2019study}, we use it specifically for forms borrowed from languages that do not share the relevant inherited Semitic root. Distinguishing these lexical relationships requires integrating orthographic, phonological, and semantic evidence rather than relying on surface similarity alone.

% Some preserve both surface form and meaning across languages, such as Hebrew \sethebrew\<ילד> (\textit{ild}, `child') and Arabic \setarab\<ولد> (\textit{wld}, `boy/child'), and are treated as True Cognates. Others retain orthographic or phonological similarity while undergoing semantic divergence, yielding False Friends (i.e., False Cognates). For example, Hebrew \sethebrew\<לחם> (\textit{lxm}, `bread') and Arabic \setarab\<لحم> (\textit{lHm}, `meat') share the Semitic root l-H-m but differ in meaning. We further distinguish these categories from loawords. Although loanwords are generally defined as words borrowed from one language into another \citep{mahajna2019study}, we use the term to refer to forms borrowed from languages that do not share the relevant inherited Semitic root. For instance, Arabic \setarab\<سوشي> (\textit{sw{\SHIN}y}, `sushi') and Hebrew \sethebrew\<סושי> (\textit{sw{\SHIN}y}, `sushi') are orthographically and phonologically similar because both are modern borrowings rather than inherited Semitic cognates. The close relationship between Arabic and Hebrew therefore creates a challenging setting for lexical reasoning, where highly similar forms may reflect shared inheritance, semantic divergence, or independent borrowing. Distinguishing among these cases requires integrating orthographic, phonological, and semantic evidence rather than relying on surface similarity alone.

\subsection{Modeling Cognate Relations}
Computational modeling of cognate relations has traditionally focused on identifying historically related words across languages. Early work relied on orthographic and phonological similarity-based methods \cite{listsca2012, list2012lexstat,hauer2015automatic}, later incorporating probabilistic, clustering, and feature-based techniques \cite{rama-2018-similarity,rama-list-2019-automated,macsween-caines-2020-expectation,cristea-etal-2021-automatic-discrimination}. More recent work has incorporated neural architectures and cross-lingual representations \cite{kanojia2020true,kanojia-etal-2020-harnessing,kanojia-etal-2021-cognition,bafna-etal-2022-combining,nath-etal-2022-phonetic,akavarapu-bhattacharya-2023-cognate,akavarapu-bhattacharya-2024-automated,uban-etal-2025-friend}.

% Cognate databases have also been assembled to support this line of work~\citep{batsuren-etal-2019-cognet,batsuren2022large}.

Alongside these methods, a range of resources has been developed for cognate identification \cite{batsuren-etal-2019-cognet,cristea-etal-2021-automatic-discrimination,batsuren2022large,list-etal-2022-sigtyp}. Two efforts are particularly close to our work. \citet{dinu-etal-2023-robocop} introduced a benchmark spanning both cognates and borrowings across Romance languages; like ours, their taxonomy treats borrowing as distinct from inheritance rather than collapsing the two. \citet{cahyawijaya2025thank} introduced StingrayBench, which evaluates multilingual LLMs on false friends in context across four language pairs, and is the closest prior benchmark to ours in both task design and evaluation setup. The former classifies lexical relations without context, whereas the latter evaluates usage in context; neither combines both settings, and neither covers Semitic languages.

Despite existing resources and models for Arabic--Hebrew processing, including machine translation, multilingual evaluation, transliteration, and cross-script representation \citep{shilon-etal-2010-machine,belinkov2016largescalemachinetranslationarabic,cettolo2016arabichebrewparallelcorpusted,zaghouani-etal-2024-fignews,gazit-etal-2025-splintering,saeed2025impactinflectionalmorphologyprobes,moreno-gonzalez-etal-2026-tale,gadd2026symphonymuniversalphoneticembeddings}, Arabic--Hebrew cognate relations remain largely unexplored computationally, and to our knowledge, no benchmark exists for evaluating them.

To address this gap, we introduce SemCog Bench, a curated  Arabic--Hebrew benchmark of true cognates, false friends, and loanwords for cognate identification and semantic disambiguation.

% Modeling Arabic and Hebrew is challenging due to their rich, complex morphology and distinct scripts, which has motivated research into morphology-aware approaches and specialized probing frameworks \citep{gazit-etal-2025-splintering,saeed2025impactinflectionalmorphologyprobes}. Prior work has developed key resources like machine translation datasets and multilingual benchmarks \citep{belinkov2016largescalemachinetranslationarabic,Cettolo2016AnAP,zaghouani-etal-2024-fignews}, while other studies have addressed script divergence through transliteration and cross-script embeddings \citep{gonzalez2026tale,gadd2026symphonymuniversalphoneticembeddings}. Broader research on multilingual representations also highlights challenges in semantic alignment, often influenced by dominant languages like English \citep{schut2025do,lim2025understanding}. Despite these foundational studies, existing work has not directly tested whether LLMs can distinguish between true cognates, false friends, and loanwords under controlled conditions. SemCog Bench is designed to fill this gap by specifically targeting cognate reasoning and false-friend disambiguation in Arabic--Hebrew.

\section{SemCog Bench}
\label{sec:semcog_bench}
\begin{figure*}[!t]
    \centering
    \includegraphics[width=\linewidth]{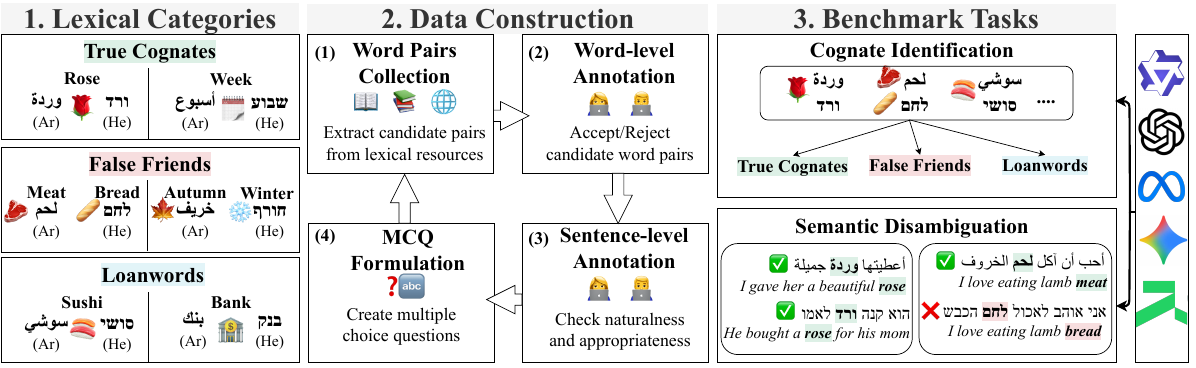}
    % \caption{Overview of the data construction pipeline and evaluation framework for SemCog Bench. (Left) Representative examples of Arabic--Hebrew true cognates, false friends, and loanwords. (Middle) The data collection process, which aggregates candidate pairs from multiple resources followed by human verification and sentence-level annotation. (Right) The downstream evaluation tasks: cognate identification and semantic disambiguation. }
    \caption{Overview of SemCog Bench, including its lexical categories, data construction, and benchmark tasks.}
    \label{fig:data_creation}
\end{figure*}

In this section, we describe the construction of SemCog Bench. Figure~\ref{fig:data_creation} summarizes the benchmark from lexical categories to task formulation. The process consists of four steps: (1) collecting candidate word pairs; (2) word-level annotation; (3) sentence-level validation; and (4) formulating multiple-choice questions for cognate identification and semantic disambiguation.

% In this section, we describe the construction of SemCog Bench. Figure~\ref{fig:data_creation} illustrates the three-stage construction pipeline: (1) cognate collection, where candidate word pairs are gathered from linguistic resources; (2) word- and sentence-level annotation, where annotators verify lexical relations and validate contextual sentences; and (3) task creation, where validated examples are converted into benchmark instances.

% In this section, we introduce the construction of a comprehensive dataset that covers cognate-related word pairs in Arabic and Hebrew. We illustrate the overall data annotation pipeline in Figure~\ref{fig:data_creation}. The  pipeline comprises three main stages. In the first stage, cognate collection, we gather candidate True Cognates (TCs), False Friends (FFs), and Loanwords (LWs) from various linguistic resources. In the second stage, word- and sentence-level annotation, human annotators evaluate the quality of the collected word pairs and assess the naturalness of the contextual sentences provided for each word. Finally, the third stage, MCQ creation, involves the formulation of the MCQs based on the validated sentences.

% https://drive.google.com/file/d/1JMaeq7p7XzPCH_gLJDA3uVSv6_rS8tsv/view?usp=drive_link

% \subsection{Categories of Word Pairs}
\subsection{Candidate Word Pairs Collection}

We collect candidate Arabic--Hebrew word pairs from multiple linguistic resources. For true cognates and false friends, we primarily rely on a comparative Arabic--Hebrew lexicon\footnote{\url{https://seveleu.com/pages/semitic-syntax-morpho/comparative-sem}} and the bilingual dictionary of the Palisra project~\cite{habash2002palisra,habash2023palisri}, which contains aligned Arabic and Hebrew word pairs.\footnote{\url{https://www.palisra.com/}}
For loanwords, we compile word candidates from Wiktionary, the Academy of the Hebrew Language materials, and prior studies on loanword phonology, morphological adaptation, and lexical borrowing; the complete list of sources is provided in Appendix~\ref{appendix:data_resources}. 

Following aggregation, we apply part-of-speech filtering and deduplication, retaining only nouns, verbs, adjectives, and adverbs, as these content words carry the semantic information required for sentence-level interpretation. This yields 1,016 true cognates, 446 false friends, and 724 loanwords. The resulting candidate lists are then reviewed by human annotators for validation. 
% We collect candidate Arabic--Hebrew word pairs from various linguistic resources, including a comparative lexicon, Hebrew and Arabic source words from a language dictionary, and loanword lists from Wikipedia alongside other resources. Detailed information regarding the data sources is provided in Appendix~\ref{appendix:data_resources}. Following the initial aggregation of these lexical items, we apply a part-of-speech (POS) filter and remove duplicate entries. We retain only nouns, verbs, adjectives, and adverbs, because these content words carry the primary semantic information necessary for sentence-level meaning. The resulting filtered candidate lists are then provided to human annotators for the formal word annotation process. This collection process yields 1,027 true cognates, 474 false friends, and 910 loanwords.

\subsection{Word- and Sentence-Level Annotation}

\paragraph{Word-Level Annotation}
To ensure dataset quality, we employ two fairly compensated annotators with native proficiency in Arabic and Hebrew and formal training in linguistics. One annotator holds an MA in Semitic languages, while the other holds a PhD in linguistics. A primary annotator reviews and accepts or rejects the initial classification of all candidate word pairs.
% In the word-level evaluation, one annotator independently verify the initial classifications, we randomly select 10\% data to calculate IAA with the second annotator.
For true cognates and false friends, the annotator assesses the presence of semantic overlap between the Arabic and Hebrew words. Pairs sharing at least one core meaning are retained as true cognates, whereas pairs with completely divergent meanings are retained as false friends.
% For loanwords, the annotators verify whether the phonologically similar pair originates from a non-Semitic source language and shares the same meaning.
For loanwords, the annotator verifies that both words originate from the same non-Semitic source language and preserve the same meaning. The annotator is also instructed to reject archaic terms, named entities such as person and location names, and lexical items that are not commonly used in both Modern Standard Arabic and Modern Hebrew.
% The validated word pairs form the basis of the cognate identification task (\S\ref{sec:evaluation_pipeline}).

\paragraph{Inter-Annotator Agreement}
To assess consistency in the word-level annotations, the second annotator independently accepted or rejected the initial classifications for two random 10\% subsets: 149 true-cognate/false-friend pairs and 84 loanword pairs. Because both annotators accepted the large majority of the proposed labels, we measure inter-annotator agreement using Gwet's AC1, which is more robust to class imbalance than Cohen's $\kappa$ and Krippendorff's $\alpha$ \cite{wongpakaran2013kappa,Ohyama03082021}. For the true-cognate/false-friend subset, raw agreement was 85\%, with a Gwet's AC1 of 0.80. For reference, Cohen's $\kappa$ and Krippendorff's $\alpha$ were both 0.32 on the same subset. The lower $\kappa$ and $\alpha$ values are consistent with the skewed marginals, as the primary and second annotators accepted 88\% and 86\% of the pairs, respectively \cite{feinstein1990}. Of the 23 disagreements (15\%), 22 involved cases where one annotator accepted the initial true-cognate or false-friend classification while the other rejected it because they considered the opposite relation more appropriate. The remaining disagreement involved a true-cognate candidate rejected by one annotator as a named entity.

Based on the primary annotator's decisions, we retain 1,858 word pairs: 858 true cognates, 364 false friends, and 636 loanwords as the validated word-level set (Table \ref{tab:dataset-stats}).

\paragraph{Sentence-Level Annotation}
Following word-level validation, we use Gemini-3.1-pro-preview \cite{gemini} to generate two contextual sentences for each validated Arabic--Hebrew word pair: one Arabic sentence containing the Arabic target word and one Hebrew sentence containing the corresponding Hebrew target word. This yields 3,716 generated sentences for the 1,858 validated word pairs. The generation prompt is provided in Appendix~\ref{apdx:sentence_generation_prompt}. One annotator then reviews all generated sentences for naturalness. A sentence is considered natural if it is grammatically correct and representative of native-speaker usage. For sentences deemed unnatural due to grammatical errors or implausible contexts, the annotator provides a revised sentence that preserves the intended meaning and target word. Of the 3,716 generated sentences, 3,549 (9.5\%) were retained unchanged, 151 (4.1\%) received minor lexical edits, and 16 (0.4\%) were substantially rewritten. This review process reduces obvious generation artifacts, including awkward or non-idiomatic phrasing.
% After the word-level filtering, the annotators evaluate the generated Arabic and Hebrew sentences based on naturalness.
Detailed annotation guidelines for both the word- and sentence-level annotation tasks are provided in Appendix~\ref{app:annotation-guidlines}.

\subsection{Benchmark Tasks}
\label{sec:evaluation_pipeline}
SemCog Bench comprises two benchmark tasks: Cognate Identification (CI) and Semantic Disambiguation (SD). Both tasks are formulated as multiple-choice questions (MCQs) designed to evaluate whether models can distinguish surface-form similarity from semantic equivalence.

% We investigate two primary experimental configurations: (i) Cognate Identification and (ii) Semantic Disambiguation. After constructing the sentence pairs, we formalize the evaluation as MCQs.

\paragraph{Cognate Identification (CI)}
Given an Arabic--Hebrew word pair $(W_{Ar}, W_{He})$, the model must determine whether the pair represents a true cognate, false friend, or loanword. This task evaluates whether models can distinguish lexical similarity arising from shared meaning, semantic divergence, or borrowing, rather than relying solely on surface-form resemblance. We consider two input settings. In the \textit{word-only} setting, the model receives only the target word pair. In the \textit{contextual} setting, the model additionally receives contextual sentences in Arabic and Hebrew, $(S_{Ar}, S_{He})$, providing contextual evidence about the meaning of each word. We evaluate both a three-class setting involving all three categories and a binary setting restricted to true cognates and false friends. The latter removes the confounding effect of loanwords and provides a more direct measure of the model's ability to resolve form--meaning conflicts.

% In this phase, we evaluate the capability of a model $L(\cdot)$ to classify a given pair of Arabic and Hebrew words $(W_{Ar}, W_{He})$ as a true cognate, a false friend, or a loanword. The model receives the target word pair, optionally accompanied by a contextual sentence $S_{Ar}$ or $S_{He}$ to ground the semantic usage of the words. By comparing the classification performance across these categories, we aim to determine whether the LLM relies primarily on surface-level morphological overlap or possesses genuine cross-lingual semantic reasoning capabilities. Additionally, we conduct a binary classification experiment restricted to true cognates and false friends. The objective of comparing this binary setting with the ternary formulation is to evaluate the tendency of the models to misclassify true cognates and false friends as loanwords when the third category is introduced. This binary setting serves as a controlled baseline to isolate the capacity of the models for pure cross-lingual semantic disambiguation, thereby allowing us to precisely quantify the confounding effect introduced by loanwords in the ternary task.

\paragraph{Semantic Disambiguation (SD)}
Given a word pair $(W_{Ar}, W_{He})$ and their corresponding contextual sentences $(S_{Ar}, S_{He})$, the model must determine which sentence or sentences are semantically correct. Each instance is formulated as a three-way MCQ with the following choices: (A) only $S_{Ar}$ is semantically correct; (B) only $S_{He}$ is semantically correct; and (C) both sentences are correct.

For true cognate and loanword pairs, we use the validated sentences collected during sentence-level annotation. Since both sentences employ words with the same meaning, they are semantically valid and the correct answer is therefore (C) (Figure \ref{fig:example_ar_he}(a,d)).

% For example, Arabic \setarab\<وردة> (\textit{wrd{\TAMARBUTA}}, `rose') and Hebrew \sethebrew\<ורד> (\textit{wrd}, `rose') form a true cognate pair; consequently, both of the following sentences are semantically valid:

% \begin{quote}
% \footnotesize
% \setarab{
% \begin{arabtext}
% أعطيتها وردة جميلة في عيد ميلادها
% \end{arabtext}}
% (I gave her a beautiful rose on her birthday)
% \par
% {\sethebrew
% \begin{arabtext}
% הוא קנה ורד אדום לאמו
% \end{arabtext}}
% (He bought a red rose for his mother)
% \end{quote}

For false friends, we start from the validated Arabic and Hebrew sentences for each false-friend pair and create minimally different sentence pairs in which one sentence is semantically correct and the other is semantically incorrect. For each direction, an annotator translates the validated sentence into the other language and replaces the correct translation of the target word with its false-friend counterpart while preserving agreement and inflection. The resulting sentence remains grammatically well-formed but becomes semantically implausible (Figure~\ref{fig:example_ar_he}(b,c)). We apply this procedure in both directions (Arabic$\leftrightarrow$Hebrew), yielding two SD instances per false-friend pair.
% False friend pairs require a different construction. To evaluate whether models rely on meaning rather than surface-form similarity, we create minimally different sentence pairs in which one sentence is semantically correct and the other is semantically incorrect. Starting from a manually validated sentence, an annotator produces a translation in the other language and replaces the correct translation of the target word with its false friend counterpart while preserving agreement and inflection. As a result, the corrupted sentence remains grammatically well-formed but becomes semantically implausible. We apply this procedure in both directions (Arabic$\leftrightarrow$Hebrew), yielding two SD instances for each false-friend pair.

% For example, \setarab\<متن> (\textit{mtn}, `back') translates to Hebrew as \sethebrew\<גב> (\textit{gb}, `back'), whereas the Hebrew word \sethebrew\<מתנו> (\textit{mtnw}, `his gift') serves as a false friend. We therefore replace the correct Hebrew form \sethebrew\<גבו> (\textit{gbw}, `his back') with the false friend {\sethebrew\<מתנו>} (\textit{mtnw}, `his gift'), yielding a Hebrew sentence that is grammatically well-formed but semantically incorrect.

% \begin{quote}
% \footnotesize
% \setarab{
% \begin{arabtext}
% كان الجندي يحمل حقيبته الثقيلة على متن ظهره طوال اليوم
% \end{arabtext}}
% (The soldier carried his heavy bag on his back all day)
% \par
% {\sethebrew
% \begin{arabtext}

% החייל סחב את התיק הכבד שלו על צד מתנו כל היום
% \end{arabtext}}
% (The soldier carried his heavy bag on his gift all day)
% \end{quote}
In total, the SD task contains 2,222 MCQ instances: 858 true cognates, 728 false friends, and 636 loanwords (Table~\ref{tab:dataset-stats}).
% The false-friend count is doubled because the bidirectional corruption procedure generates two MCQs per pair.

% \vspace{-1em}

% For both experimental configurations, we employ diverse transcription and transliteration formats to present the Arabic and Hebrew inputs. Specifically, the experiments are conducted on the original scripts, Romanized transliterated texts, transcribed phonetic strings (IPA), or combinations thereof. These systematic variations assist in investigating whether the evaluated LLMs process cross-lingual cognates phonetically or predominantly rely on script-specific orthographic features.

\begin{table}[t]
\centering
% \footnotesize
\setlength{\tabcolsep}{5pt}
\begin{tabular}{lcccc}
\toprule
\textbf{Type} & \textbf{\#Pairs} & \textbf{Ar Len.} & \textbf{He Len.} & \textbf{\#MCQs} \\
\midrule
TC    &   858 & 13.0 & 12.8 &   858 \\
FF    &   364 & 13.1 & 12.8 &   728 \\
LW    &   636 & 14.9 & 14.4 &   636 \\
\midrule
Total & 1,858 & 13.7 & 13.4 & 2,222 \\
\bottomrule
\end{tabular}
\caption{Statistics of SemCog Bench. TC, FF, and LW denote true cognates, false friends, and loanwords, respectively. Ar Len. and He Len. are the average lengths (in words) of the Arabic and Hebrew sentences.}
% \caption{Statistics of SemCog Bench. TC, FF, and LW denote true cognates, false friends, and loanwords, respectively. Sentence lengths are reported as average word counts for Arabic and Hebrew sentences. For false friends, \#MCQs doubled because each word pair yields two instances, each constructed by pairing one valid sentence with a corrupted sentence as the incorrect option.}
\label{tab:dataset-stats}
\end{table}

\begin{table*}[!t]
    \centering
    \small
    \setlength{\tabcolsep}{4pt}
    \renewcommand{\arraystretch}{1.2}
    
    \definecolor{HeaderBg}{RGB}{235,240,248}
    \definecolor{PinkBg}{RGB}{252, 228, 236}
    \definecolor{GreenBg}{RGB}{232, 245, 233}
    \definecolor{GrayBg}{RGB}{245, 245, 245}
    \definecolor{GrayBgDark}{RGB}{240,240,240}

    \resizebox{\textwidth}{!}{%
        \begin{tabular}{l ccccccc ccccccc}
            \toprule
            & \multicolumn{7}{c}{\textbf{Cognate Identification (CI)}} & \multicolumn{7}{c}{\textbf{Semantic Disambiguation (SD)}} \\
            \cmidrule(lr){2-8} \cmidrule(lr){9-15}
            \textbf{Model} & \textbf{Acc\textsubscript{TC}} & \textbf{Acc\textsubscript{FF}} & \textbf{Acc\textsubscript{LW}} & \textbf{Acc} & \textbf{P} & \textbf{R} & \textbf{F\textsubscript{1}}  & \textbf{Acc\textsubscript{TC}} & \textbf{Acc\textsubscript{FF}} & \textbf{Acc\textsubscript{LW}} & \textbf{Acc} & \textbf{P} & \textbf{R} & \textbf{F\textsubscript{1}} \\
            \midrule

            \rowcolor{GrayBgDark}
            \multicolumn{15}{c}{\textbf{Proprietary LLMs}} \\
            GPT-4o                & 62.5 & 71.4 & 90.7 & 73.9 & 73.5 & 74.9 & 72.5 & \textbf{94.6} & 35.6 & 93.1 & 74.8 & 75.2 & 74.2 & 73.7 \\
            GPT-5.4               & 85.9 & 47.0 & \textbf{96.4} & 81.9 & 78.3 & 76.4 & 76.9 & 88.7 & 57.7 & 90.2 & 79.0 & 79.0 & 79.1 & 78.9 \\
            DeepSeek-v4-pro           & 91.6 &  1.1 & 83.5 & 71.1 & 65.6 & 58.7 & 55.8 & 17.0 & 50.1 & 19.3 & 28.5 & 25.3 & 28.3 & 24.0 \\
            Qwen3.6-Plus          & 87.3 & 47.8 & 94.8 & \textbf{82.1} & \textbf{78.6} & \textbf{76.6} & \textbf{77.3} & 91.5 & 50.8 & \textbf{94.3} & 79.0 & 79.1 & 78.9 & 78.9 \\

            \midrule
            \rowcolor{GrayBgDark}
            \multicolumn{15}{c}{\textbf{Open-source LLMs}} \\
            Qwen2.5-7B-Instruct   & 64.9 & 64.6 & 25.0 & 51.2 & 63.2 & 51.5 & 47.9 & 43.1 & 40.7 & 40.1 & 41.4 & 37.1 & 41.1 & 36.6 \\
            Qwen3-8B              & 75.2 & 52.7 & 18.6 & 51.4 & 63.0 & 48.8 & 45.1 & 87.8 &  6.3 & 84.1 & 60.0 & 61.7 & 59.5 & 58.6 \\
            Llama-3.1-8B-Instruct & 91.7 & 12.1 &  0.0 & 44.7 & 22.3 & 34.6 & 27.1 & 31.9 & 30.2 & 41.3 & 34.1 & 39.8 & 35.1 & 29.8 \\
            Llama-3.3-70B-Instruct & 87.9 & 23.1 & 92.9 & 76.9 & 71.8 & 68.0 & 67.5 & 80.1 &  8.1 & 82.7 & 57.2 & 57.8 & 56.9 & 56.5 \\
            GLM-4-9B-Chat         & 88.0 & 24.7 & 29.2 & 55.5 & 60.4 & 47.3 & 47.1 & 26.7 & 52.9 & 30.3 & 36.3 & 23.2 & 36.1 & 26.2 \\
            Gemma-2-9B-IT         & 80.7 & 23.9 & 66.2 & 64.6 & 60.9 & 56.9 & 57.1 & 38.1 & 13.9 & 41.2 & 31.0 & 42.3 & 30.9 & 22.5 \\
            Gemma-4-31B-IT         & 84.0 & 38.7 & 93.9 & 78.5 & 74.1 & 72.2 & 72.5 & 90.7 & \textbf{58.0} & 89.3 & \textbf{79.6} & \textbf{79.6} & \textbf{79.6} & \textbf{79.5} \\
            Jais-Adapted-7B-Chat  & 62.4 &  2.2 & 36.9 & 41.9 & 41.7 & 33.8 & 31.0 & 33.7 & 30.6 & 34.6 & 32.9 & 33.8 & 33.5 & 18.9 \\
            Jais-2-8B-Chat        & \textbf{96.2} &  8.0 &  0.0 & 46.0 & 26.0 & 34.7 & 25.2 & 33.8 & 29.7 & 33.3 & 32.3 & 13.3 & 32.9 & 16.4 \\
            DictaLM-2.0-Instruct  & 33.3 & \textbf{82.4} &  0.2 & 31.6 & 48.3 & 38.6 & 26.4 & 34.2 & 29.5 & 34.4 & 32.7 & 10.9 & 33.3 & 16.4 \\
            Aya-23-8B             &  6.3 & 39.0 & 90.3 & 41.4 & 47.8 & 45.2 & 34.7 & 38.0 & 26.7 & 44.6 & 36.2 & 55.1 & 37.0 & 25.2 \\
            GPT-OSS-120B          & 66.6 & 54.4 & 82.7 & 69.7 & 70.0 & 71.2 & 69.8 & 47.0 & 12.2 & 47.3 & 35.7 & 53.7 & 50.3 & 46.9 \\

            \bottomrule
        \end{tabular}
    }%
    % \caption{Main results on Cognate Identification (3-class) and Semantic Disambiguation tasks. All values are percentages (\%). P, R, and F\textsubscript{1} are macro-averaged across TC, FF, and LW classes. For CI, macro P/R/F\textsubscript{1} are computed from confusion matrices; for SD, macro P/R/F\textsubscript{1} are computed from per-answer-option (A/B/C) confusion matrices within each sub-experiment, then macro-averaged across TC, FF, and LW sub-experiments.}
    \caption{Results on three-class cognate identification (CI) and semantic disambiguation (SD) using undiacritized inputs. We report overall and class-specific accuracy (Acc), as well as macro-averaged precision (P), recall (R), and F\textsubscript{1}. TC, FF, and LW denote true cognates, false friends, and loanwords, respectively. The best result in each column is shown in \textbf{bold}.}
    \label{tab:main_results}
\end{table*}

% \subsection{Dataset Statistics}

% For each validated pair, we retain the target words and their corresponding contextual sentences, denoted as $(W_{Ar}, S_{Ar})$ and $W_{He}, S_{He})$. In terms of the IAA, the two annotators achieve an initial raw agreement of 92\%. Most disagreements involve borderline cases between potential loanwords and true cognates, which motivates an additional verification round. After filtering potential loanwords from the true cognate category, inter-annotator agreement reaches 0.87 for TC/FF and 0.81 for LW under Gwet's AC1, indicating substantial agreement.

% Krippendorff's alpha scores are lower, at 0.38 and 0.45, respectively. We interpret these values cautiously, as Krippendorff's alpha can be sensitive to highly imbalanced label distributions and may underestimate agreement when one label dominates~\cite{vach2023gwet}. This is relevant in our setting, where accept decisions are much more frequent than reject decisions. Overall, the high raw agreement, substantial Gwet's AC1 scores, and additional verification step support the reliability of the annotations. Dataset statistics are reported in Table~\ref{tab:dataset-stats}.

\section{Experimental Setup}
\label{sec:experiment}
We evaluate a diverse set of LLMs on SemCog Bench and examine the effects of input representation, sentence-level context, and model scale.

\paragraph{LLMs}
We evaluate LLMs categorized into four groups: (1) \textbf{Multilingual Models}, including Qwen2.5-7B-Instruct~\cite{qwen2025qwen25technicalreport}, Qwen3-8B~\cite{yang2025qwen3technicalreport}, GLM-4-9B-Chat~\cite{glm2024chatglmfamilylargelanguage}, Llama-3.1-8B-Instruct and Llama-3.3-70B-Instruct~\cite{grattafiori2024llama3herdmodels}, Gemma-2-9B-IT and Gemma-4-31B-IT~\cite{gemmateam2024gemma2improvingopen}, Aya-23-8B~\cite{aryabumi2024aya23openweight}, and GPT-OSS-120B  \cite{openai2025gptoss}; (2) \textbf{Arabic-centric Models}, comprising Jais-2-8B-Chat \cite{anwar2026jais2familyarabiccentric} and Jais-Adapted-7B-Chat~\cite{sengupta2023jais}; (3) \textbf{Hebrew-capable Models}, represented by DictaLM-2.0-Instruct~\cite{shmidman2024adaptingllmshebrewunveiling}; and (4) \textbf{Proprietary Models}, which include GPT-5.4 \cite{gpt5}, GPT-4o \cite{openai2024gpt4technicalreport}, DeepSeek-v4-pro \cite{deepseekai2025deepseekv3technicalreport}, and Qwen3.6-Plus \cite{yang2025qwen3technicalreport}. Proprietary LLM snapshots and licenses for all models are provided in Appendix~\ref{apdx:model_snapshots}, hyperparameters in Appendix~\ref{apdx:experiment_settings}, and prompts in Appendix~\ref{appendix:evaluation_prompts}.

\begin{table}[th]
\centering
% \small
\setlength{\tabcolsep}{4pt}
\begin{tabular}{lcccc}
\toprule
\textbf{Model} &
\textbf{Acc\textsubscript{TC}}&
\textbf{Acc\textsubscript{FF}} &
\textbf{Acc} &
\textbf{F\textsubscript{1}} \\
\midrule
LingPy       & 79.1 & 23.4 & 62.5 & 50.9 \\
FastText  & 61.8 & \textbf{61.5} & 61.7 & 59.1 \\
% FastText JSD     & 58.9 & 54.4 & 57.5 & 54.7 \\
SBERT     & 54.9 & \textbf{61.5} & 56.9 & 55.0 \\
% SBERT JSD        & 53.7 & 59.3 & 55.4 & 53.5 \\
\midrule
Qwen3.6-Plus          & \textbf{94.1} & 35.7 & \textbf{76.7} & 66.3 \\
Gemma-4-31B-IT        & 88.9 & 43.1 & 75.3 & \textbf{68.4} \\
% GPT-5.4          & \textbf{85.9} & 47.0 & \textbf{74.3} & \textbf{69.4} \\
\bottomrule
\end{tabular}
\caption{Results on binary cognate identification over 1,222 true-cognate/false-friend (TC/FF) word pairs in terms of accuracy (Acc) and F\textsubscript{1}.}
\label{tab:simple_baselines}
\end{table}

\begin{table*}[t]
\definecolor{GrayBgDark}{RGB}{240,240,240}
\centering
\small
\setlength{\tabcolsep}{2.5pt}
\renewcommand{\arraystretch}{1.2}
\begin{tabular}{l rrrrrr rrrrrr}
\toprule
 & \multicolumn{6}{c}{\textbf{Cognate Identification (CI)}} & \multicolumn{6}{c}{\textbf{Semantic Disambiguation (SD)}} \\
\cmidrule(lr){2-7}\cmidrule(lr){8-13}
 & \multicolumn{2}{c}{\textbf{Diacritized}} & \multicolumn{2}{c}{\textbf{IPA}} & \multicolumn{2}{c}{\textbf{Romanized}} & \multicolumn{2}{c}{\textbf{Diacritized}} & \multicolumn{2}{c}{\textbf{IPA}} & \multicolumn{2}{c}{\textbf{Romanized}} \\
\cmidrule(lr){2-3}\cmidrule(lr){4-5}\cmidrule(lr){6-7}\cmidrule(lr){8-9}\cmidrule(lr){10-11}\cmidrule(lr){12-13}
\textbf{Model} & \multicolumn{1}{c}{\textbf{Acc}} & \multicolumn{1}{c}{\textbf{F\textsubscript{1}}} & \multicolumn{1}{c}{\textbf{Acc}} & \multicolumn{1}{c}{\textbf{F\textsubscript{1}}} & \multicolumn{1}{c}{\textbf{Acc}} & \multicolumn{1}{c}{\textbf{F\textsubscript{1}}} & \multicolumn{1}{c}{\textbf{Acc}} & \multicolumn{1}{c}{\textbf{F\textsubscript{1}}} & \multicolumn{1}{c}{\textbf{Acc}} & \multicolumn{1}{c}{\textbf{F\textsubscript{1}}} & \multicolumn{1}{c}{\textbf{Acc}} & \multicolumn{1}{c}{\textbf{F\textsubscript{1}}} \\
\midrule
\rowcolor{GrayBgDark}
\multicolumn{13}{c}{\textbf{Proprietary LLMs}} \\
GPT-4o & \D{+1.7} & \D{+0.6} & \Ds{-15.1} & \Ds{-13.0} & \Ds{-32.1} & \Ds{-30.5} & \Ds{-29.9} & \Ds{-29.4} & \Ds{-45.7} & \Ds{-46.2} & \Ds{-45.5} & \Ds{-47.3} \\
GPT-5.4 & \Ds{-1.8} & \Ds{-2.7} & \Ds{-11.3} & \Ds{-9.5} & \Ds{-20.1} & \Ds{-18.6} & \Ds{-45.9} & \Ds{-46.0} & \Ds{-43.8} & \Ds{-46.1} & \Ds{-54.7} & \Ds{-56.0} \\
DeepSeek-v4-pro & \Ds{-3.0} & \Ds{-3.9} & \Ds{-10.8} & \Ds{-11.9} & \Ds{-17.5} & \Ds{-19.9} & \Ds{-9.4} & \Ds{-7.5} & \Ds{-10.5} & \Ds{-6.8} & \Ds{+3.7} & \Ds{-6.1} \\
Qwen3.6-Plus & \D{-2.7} & \D{-3.3} & \D{-16.0} & \D{-15.3} & \Ds{-30.8} & \Ds{-28.3} & \D{-44.1} & \D{-45.0} & \D{-42.7} & \D{-45.6} & \Ds{-53.8} & \Ds{-55.1} \\
\midrule
\rowcolor{GrayBgDark}
\multicolumn{13}{c}{\textbf{Open-source LLMs}} \\
Qwen2.5-7B-Instruct & \Ds{-2.6} & \Ds{-5.6} & \Ds{-25.8} & \Ds{-25.2} & \Ds{-21.0} & \Ds{-20.5} & \Ds{-8.5} & \Ds{-4.5} & \Ds{-12.6} & \Ds{-5.5} & \Ds{-20.3} & \Ds{-17.5} \\
Qwen3-8B & \Ds{-3.0} & \Ds{-7.0} & \Ds{-22.1} & \Ds{-20.4} & \Ds{-27.6} & \Ds{-26.2} & \Ds{-8.4} & \Ds{-12.9} & \Ds{+3.2} & \Ds{+3.6} & \Ds{-30.5} & \Ds{-37.3} \\
Llama-3.1-8B-Instruct & \D{+0.9} & \Ds{-1.5} & \D{-0.6} & \D{-1.0} & \Ds{-10.0} & \Ds{+5.3} & \Ds{-3.0} & \Ds{-3.8} & \D{+1.9} & \Ds{-5.3} & \Ds{+12.1} & \Ds{-6.6} \\
Llama-3.3-70B-Instruct & \Ds{-7.1} & \Ds{-6.0} & \Ds{-32.2} & \Ds{-26.9} & \Ds{-33.0} & \Ds{-26.3} & \Ds{-2.0} & \Ds{-3.2} & \Ds{-5.4} & \Ds{-5.3} & \Ds{-17.8} & \Ds{-19.2} \\
GLM-4-9B-Chat & \Ds{-2.9} & \Ds{-5.4} & \Ds{-29.1} & \Ds{-22.6} & \Ds{-27.7} & \Ds{-18.6} & \D{-1.5} & \Ds{-6.0} & \Ds{-3.0} & \Ds{-9.6} & \D{-0.9} & \Ds{-18.3} \\
Gemma-2-9B-IT & \D{-1.1} & \Ds{-5.2} & \Ds{-24.2} & \Ds{-26.2} & \Ds{-16.0} & \Ds{-21.1} & \Ds{-1.9} & \Ds{-5.5} & \Ds{+1.6} & \Ds{-6.1} & \Ds{-10.9} & \Ds{-16.7} \\
Gemma-4-31B-IT & \D{-0.7} & \Ds{-2.9} & \Ds{-11.9} & \Ds{-9.4} & \Ds{-21.1} & \Ds{-18.4} & \Ds{-43.1} & \Ds{-42.5} & \Ds{-46.9} & \Ds{-46.4} & \Ds{-54.7} & \Ds{-54.3} \\
Jais-Adapted-7B-Chat & \Ds{+3.8} & \Ds{-3.7} & \Ds{-4.1} & \Ds{-3.3} & \Ds{-7.0} & \Ds{-12.8} & \D{-0.2} & \Ds{-2.5} & \D{-0.2} & \Ds{-2.5} & \Ds{-13.5} & \Ds{-21.0} \\
Jais-2-8B-Chat & \D{-0.4} & \Ds{-2.2} & \Ds{-5.4} & \D{+1.2} & \Ds{-17.2} & \D{+3.1} & \Ds{-0.8} & \D{+0.3} & \D{-0.5} & \Ds{+1.0} & \Ds{+13.2} & \Ds{+7.8} \\
DictaLM-2.0-Instruct & \D{+1.2} & \Ds{+5.5} & \Ds{-12.0} & \Ds{-15.4} & \Ds{-12.0} & \Ds{-15.4} & \D{0.0} & \D{0.0} & \D{0.0} & \D{0.0} & \D{0.0} & \D{0.0} \\
Aya-23-8B & \D{+0.2} & \D{-0.3} & \Ds{-7.8} & \Ds{-4.0} & \Ds{-7.5} & \D{-1.4} & \Ds{+3.5} & \Ds{+7.3} & \Ds{-3.5} & \Ds{-8.8} & \Ds{+16.3} & \Ds{+10.3} \\
GPT-OSS-120B & \Ds{-5.8} & \Ds{-5.7} & \Ds{-24.8} & \Ds{-20.1} & \Ds{-30.2} & \Ds{-27.3} & \Ds{-20.0} & \Ds{-20.9} & \Ds{-14.6} & \Ds{-19.3} & \Ds{-5.4} & \Ds{-7.0} \\
\bottomrule
\end{tabular}
\caption{Change ($\Delta$) in accuracy (Acc) and F\textsubscript{1} relative to the undiacritized input across different input representations.  Cell shading encodes the magnitude and direction of change, from \textcolor{DeltaNeg}{\textbf{red}} (degradation) to \textcolor{DeltaPos}{\textbf{green}} (improvement). Asterisks (*) indicate statistically significant differences ($p < 0.05$).}
\label{tab:repr_relative_changes}
\end{table*}

% \begin{figure*}[t]
%     \centering
%     \includegraphics[width=\textwidth]{figures/Analysis_relative_changes_representation_heatmap_new.pdf}
%     % \caption{Relative performance changes ($\Delta$) compared to the undiacritized baseline under different input representations. All values represent percentage point differences. $\Delta$Acc denotes the change in weighted accuracy, and $\Delta$F$_1$ denotes the change in average F\textsubscript{1} scores.}
%     \caption{Change ($\Delta$) in accuracy (Acc) and average F\textsubscript{1} relative to the undiacritized baseline across different input representations. Values are reported in percentage points. Asterisks (*) indicate statistically significant differences ($p<0.05$).}
%     % \vspace{-1em}
%     \label{fig:relative_changes_representation}
% \end{figure*}

\paragraph{Baselines} We include simple non-generative baselines to contextualize LLM performance. For Cognate Identification (CI), we evaluate LingPy's Sound-Class-Based Phonetic Alignment (SCA) algorithm \cite{list-moran-2013-open}, which operates on phonetic transcriptions (IPA), as well as embedding-based approaches following \citet{uban-etal-2025-friend}. The latter use aligned FastText embeddings \cite{grave-etal-2018-learning} and multilingual Sentence-BERT \cite{reimers-gurevych-2019-sentence} (\texttt{distiluse-base-multilingual-cased-v2}) representations, with cosine distance as the similarity measure. Because these methods distinguish only true cognates from false friends, we evaluate them on the 1,222 true-cognate/false-friend pairs. We do not include analogous baselines for Semantic Disambiguation (SD), as these similarity-based methods cannot assess whether a target word is semantically appropriate in context.
% , while perplexity-based scoring would largely overlap with the language-modeling capabilities already evaluated by the LLMs.

\paragraph{Input Representations}  To examine the effect of script and phonological form, we evaluate four input representations: the original undiacritized text, diacritized text, IPA transcriptions, and Romanized text. For Arabic, diacritics are obtained using CAMeL Tools~\cite{obeid-etal-2020-camel}, and IPA transcriptions are derived from the diacritized forms following the approach of~\citet{khalifa-etal-2025-learning} with additional linguistic mapping rules. For Hebrew, \textit{nikkud} and IPA transcriptions are generated using Phonikud~\cite{kolani2025phonikud}. Romanized forms for both languages are obtained using Uroman~\cite{hermjakob-etal-2018-box}.
% These representations allow us to examine how orthographic and phonological form affect cross-lingual lexical-semantic reasoning.

% We compare the undiacritized baseline with three alternative representations: diacritized text, IPA, and Uroman. Arabic diacritics are restored using CAMeL Tools~\cite{obeid-etal-2020-camel}, and Arabic IPA is derived from diacritized text using the method of~\citet{khalifa-etal-2025-learning} with additional linguistic mapping rules. For Hebrew, we generate \textit{nikkud} and IPA using Phonikud~\cite{kolani2025phonikud}. Uroman is used to obtain Romanized forms for both words and sentences~\cite{hermjakob-etal-2018-box}.

\paragraph{Evaluation Metrics}
For both CI and SD, we report accuracy as the primary evaluation metric. We additionally report macro-averaged F\textsubscript{1} to account for class imbalance. We also report class-specific accuracy to characterize performance across true cognates, false friends, and loanwords.

\section{Results and Analysis}

\paragraph{\colorbox{RQOneBg}{RQ1:} Identification and Disambiguation}
Table~\ref{tab:main_results} presents LLM performance on three-class CI and SD using undiacritized inputs. On CI, Qwen3.6-Plus is the strongest proprietary and overall model, achieving 82.1 accuracy and 77.3 macro F\textsubscript{1}, while Gemma-4-31B-IT is the strongest open-source model (78.5 accuracy; 72.5 F\textsubscript{1}). On SD, Gemma-4-31B-IT achieves the best overall performance (79.6 accuracy; 79.5 F\textsubscript{1}), slightly outperforming the strongest proprietary models, GPT-5.4 and Qwen3.6-Plus, which both reach 79.0 accuracy and 78.9 F\textsubscript{1}. Across both tasks, however, performance varies substantially by lexical category.

In CI, most models perform much better on true cognates than false friends; GPT-4o, DictaLM-2.0-Instruct, and Aya-23-8B show the opposite pattern, while Qwen2.5-7B-Instruct is nearly balanced. Jais-2-8B-Chat provides an extreme example, reaching 96.2 accuracy on true cognates but only 8 on false friends, consistent with a tendency to associate form similarity with shared meaning. Loanwords differ: they are the best-performing category for several models (e.g., GPT-4o, GPT-5.4, Qwen3.6-Plus), yet accuracy falls to 0 for Llama-3.1-8B-Instruct and Jais-2-8B-Chat, highlighting substantial variation in recognizing parallel borrowings.

% Loanword performance also varies widely across models, ranging from 0 accuracy for Llama-3.1-8B-Instruct and Jais-2-8B-Chat to 96.4 for GPT-5.4.

SD exhibits a more model-dependent class pattern. GPT-4o reverses its CI behavior, shifting from higher false-friend accuracy in CI (71.4 vs.\ 62.5) to substantially higher true-cognate accuracy in SD (94.6 vs.\ 35.6). DeepSeek-v4-pro and GLM-4-9B-Chat reverse in the opposite direction, performing better on false friends than true cognates in SD. DictaLM-2.0-Instruct and Aya-23-8B similarly shift from higher false-friend accuracy in CI to higher true-cognate accuracy in SD. Llama-3.1-8B-Instruct also drops sharply on true cognates, from 91.7 in CI to 31.9 in SD, resulting in nearly equal performance on true cognates and false friends. These reversals show that class-specific strengths in CI do not necessarily carry over to SD.

% A similar category-level disparity appears in SD, where false friends remain particularly challenging for many otherwise strong models.
% Overall, these results reveal a persistent difficulty in resolving cases where similar Arabic--Hebrew forms correspond to different meanings.

Table~\ref{tab:simple_baselines} compares the simple baselines with Qwen3.6-Plus and Gemma-4-31B-IT, the strongest LLMs, under the same binary CI setting. LingPy performs well on true cognates (79.1) but poorly on false friends (23.4), highlighting the limitations of relying primarily on phonetic similarity. The embedding-based baselines are more balanced across the two classes, with FastText achieving the strongest baseline performance. The LLMs achieve substantially higher overall accuracy, with Qwen3.6-Plus reaching 76.7, while Gemma-4-31B-IT obtains the highest macro F\textsubscript{1} at 68.4. Interestingly, both FastText and SBERT outperform the LLMs on false friends, whereas the LLMs are substantially stronger on true cognates. This suggests that LLMs  over-rely on form similarity when distinguishing true cognates from false friends.

\paragraph{\colorbox{RQTwoBg}{RQ2:} Input Representations}
Table~\ref{tab:repr_relative_changes} presents the change in accuracy and F\textsubscript{1} for each input representation relative to the corresponding undiacritized result; significance is assessed using a two-sided approximate randomization test with 10,000 trials \cite{dror-etal-2018-hitchhikers}. For CI, diacritization has relatively little effect, producing small gains for some models and modest degradations for others. IPA and Romanization, by contrast, generally reduce performance, often substantially. For example, Llama-3.3-70B-Instruct drops by 32.2 accuracy points and 26.9 F\textsubscript{1} under IPA, and by 33 and 26.3 points, respectively, under Romanization.

% This pattern suggests that models benefit from the original Arabic and Hebrew scripts rather than from explicitly phonetic or cross-script representations.

For SD, the effect of representation is more model-dependent. The strongest models under undiacritized input often degrade substantially when the representation is changed: GPT-5.4, Qwen3.6-Plus, and Gemma-4-31B-IT lose more than 40 accuracy points in several settings. In contrast, some lower-performing models benefit from particular representations. IPA improves both accuracy and F\textsubscript{1} for Qwen3-8B, while Romanization improves Jais-2-8B-Chat and Aya-23-8B; notably, Aya-23-8B gains 16.3 accuracy points and 10.3 F\textsubscript{1}. These results suggest that alternative representations provide no consistent advantage over undiacritized input, with their effects depending strongly on both the task and model.

\begin{table*}[t]
\definecolor{GrayBgDark}{RGB}{240,240,240}
\centering
\small
\setlength{\tabcolsep}{4pt}
\renewcommand{\arraystretch}{1.2}
\begin{tabular}{l rrrr rrrr}
\toprule
 & \multicolumn{2}{c}{\textbf{Undiacritized}} & \multicolumn{2}{c}{\textbf{Diacritized}} & \multicolumn{2}{c}{\textbf{IPA}} & \multicolumn{2}{c}{\textbf{Romanized}} \\
\cmidrule(lr){2-3}\cmidrule(lr){4-5}\cmidrule(lr){6-7}\cmidrule(lr){8-9}
\textbf{Model} & \multicolumn{1}{c}{\textbf{Acc}} & \multicolumn{1}{c}{\textbf{F\textsubscript{1}}} & \multicolumn{1}{c}{\textbf{Acc}} & \multicolumn{1}{c}{\textbf{F\textsubscript{1}}} & \multicolumn{1}{c}{\textbf{Acc}} & \multicolumn{1}{c}{\textbf{F\textsubscript{1}}} & \multicolumn{1}{c}{\textbf{Acc}} & \multicolumn{1}{c}{\textbf{F\textsubscript{1}}} \\
\midrule
\rowcolor{GrayBgDark}
\multicolumn{9}{c}{\textbf{Proprietary LLMs}} \\
GPT-4o & \Ds{+15.1} & \Ds{+15.6} & \Ds{+14.2} & \Ds{+15.5} & \Ds{+18.2} & \Ds{+16.5} & \Ds{+17.9} & \Ds{+17.9} \\
GPT-5.4 & \Ds{+10.1} & \Ds{+13.5} & \Ds{+9.9} & \Ds{+13.8} & \Ds{+16.1} & \Ds{+17.1} & \Ds{+18.4} & \Ds{+20.5} \\
DeepSeek-v4-pro & \Ds{+2.1} & \Ds{+2.2} & \Ds{-9.3} & \Ds{-2.7} & \Ds{-10.2} & \Ds{-4.3} & \D{-1.6} & \Ds{+4.5} \\
Qwen3.6-Plus & \Ds{+7.5} & \Ds{+10.8} & \Ds{+9.0} & \Ds{+13.4} & \Ds{+10.4} & \Ds{+9.0} & \Ds{+14.3} & \Ds{+12.8} \\
\midrule
\rowcolor{GrayBgDark}
\multicolumn{9}{c}{\textbf{Open-source LLMs}} \\
Qwen2.5-7B-Instruct & \Ds{+13.0} & \Ds{+14.2} & \Ds{+17.4} & \Ds{+22.0} & \Ds{+21.3} & \Ds{+24.6} & \Ds{+18.4} & \Ds{+21.3} \\
Qwen3-8B & \D{+0.6} & \Ds{-3.4} & \D{+1.1} & \Ds{-2.3} & \Ds{+7.2} & \Ds{+5.5} & \Ds{+11.4} & \Ds{+5.2} \\
Llama-3.1-8B-Instruct & \Ds{+2.0} & \Ds{-4.7} & \Ds{-1.1} & \Ds{-4.2} & \Ds{-34.0} & \Ds{-16.3} & \Ds{-3.7} & \Ds{-5.0} \\
Llama-3.3-70B-Instruct & \Ds{+5.1} & \Ds{+7.1} & \Ds{+7.1} & \Ds{+4.3} & \Ds{+11.1} & \Ds{+7.7} & \Ds{+16.5} & \Ds{+7.3} \\
GLM-4-9B-Chat & \D{-2.4} & \D{-0.3} & \D{-2.6} & \Ds{+5.5} & \Ds{+3.7} & \Ds{+5.8} & \Ds{-3.4} & \Ds{-4.0} \\
Gemma-2-9B-IT & \Ds{-7.1} & \Ds{-10.4} & \Ds{-3.2} & \Ds{-5.2} & \Ds{+4.4} & \Ds{+4.3} & \Ds{+3.9} & \D{+1.5} \\
Gemma-4-31B-IT & \Ds{+12.5} & \Ds{+17.4} & \Ds{+12.3} & \Ds{+19.0} & \Ds{+13.3} & \Ds{+15.0} & \Ds{+12.2} & \Ds{+14.8} \\
Jais-Adapted-7B-Chat & \Ds{+5.1} & \D{-1.9} & \D{-1.1} & \Ds{-4.8} & \Ds{-3.7} & \Ds{-10.6} & \Ds{+7.5} & \Ds{+9.7} \\
Jais-2-8B-Chat & \Ds{-5.5} & \Ds{+5.8} & \Ds{-5.0} & \Ds{+9.0} & \Ds{-5.0} & \D{-0.8} & \Ds{-6.3} & \Ds{-5.3} \\
DictaLM-2.0-Instruct & \Ds{+11.1} & \Ds{+3.4} & \Ds{+12.8} & \Ds{-7.7} & \Ds{+19.9} & \Ds{+15.1} & \Ds{+6.4} & \Ds{+9.5} \\
Aya-23-8B & \Ds{-13.1} & \Ds{-11.6} & \Ds{-6.8} & \D{-2.4} & \Ds{-12.9} & \Ds{-16.8} & \Ds{-6.1} & \D{-0.2} \\
GPT-OSS-120B & \Ds{+11.2} & \Ds{+12.0} & \Ds{+8.6} & \Ds{+9.0} & \D{+0.2} & \D{+2.4} & \Ds{+5.5} & \Ds{+7.3} \\
\bottomrule
\end{tabular}
\caption{Change ($\Delta$) in accuracy (Acc) and F\textsubscript{1} from word-level to sentence-level cognate identification in the three-class setting. Cell shading indicates the direction and magnitude of change, ranging from \textcolor{DeltaNeg}{\textbf{red}} (degradation) to \textcolor{DeltaPos}{\textbf{green}} (improvement). Asterisks (*) denote statistically significant differences ($p < 0.05$).}
\label{tab:sentence_context}
\end{table*}

\paragraph{\colorbox{RQThreeBg}{RQ3:} Context Effect on Cognate Identification} 
Table~\ref{tab:sentence_context} presents the change in accuracy and F\textsubscript{1} from word-level to sentence-level CI across input representations. Among proprietary models, GPT-4o, GPT-5.4, and Qwen3.6-Plus improve across all representations, with larger gains for GPT-4o and GPT-5.4. Several open-source models show consistent gains, including Qwen2.5-7B-Instruct, Llama-3.3-70B-Instruct, and Gemma-4-31B-IT.
% Table~\ref{tab:sentence_context} presents the change in accuracy and F\textsubscript{1} from word-level to sentence-level CI across input representations. Three of the four proprietary models, GPT-4o, GPT-5.4, and Qwen3.6-Plus, improve in both accuracy and F\textsubscript{1} across all representations. Qwen2.5-7B-Instruct shows similarly consistent gains, including improvements of more than 20 points under IPA.

Some models benefit particularly strongly from context under alternative input representations. DictaLM-2.0-Instruct gains 19.9 accuracy points and 15.1 F\textsubscript{1} under IPA, while Qwen2.5-7B-Instruct also achieves its largest gains under IPA (+21.3 accuracy; +24.6 F\textsubscript{1}). Other models exhibit representation-dependent effects; for example, Gemma-2-9B-IT degrades under undiacritized and diacritized inputs but improves under IPA and Romanization. Aya-23-8B, by contrast, degrades across all representations. Overall, this suggests that the ability to exploit sentence-level context for CI varies substantially across models and interacts strongly with input representation.

\begin{figure}[!t]
    \centering
    \includegraphics[width=\linewidth]{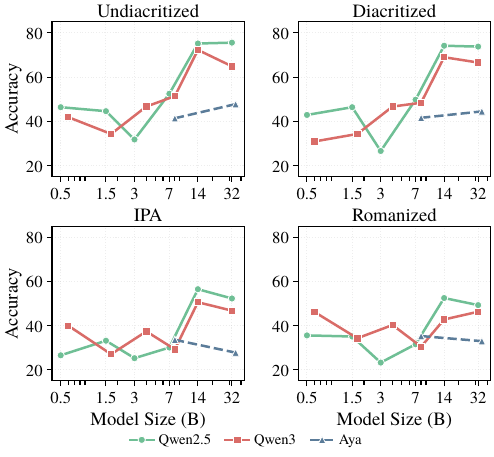}
    % \caption{Overall accuracy under different input representations for the Qwen2.5, Qwen3, and Aya models.}
    \caption{Effect of model scale on cognate identification accuracy across different input representations for the Qwen2.5, Qwen3, and Aya-23 model families.}
    % \vspace{-1em}
    \label{fig:scaling_performance_ci}
\end{figure}

\begin{figure}[!t]
    \centering
    \includegraphics[width=\linewidth]{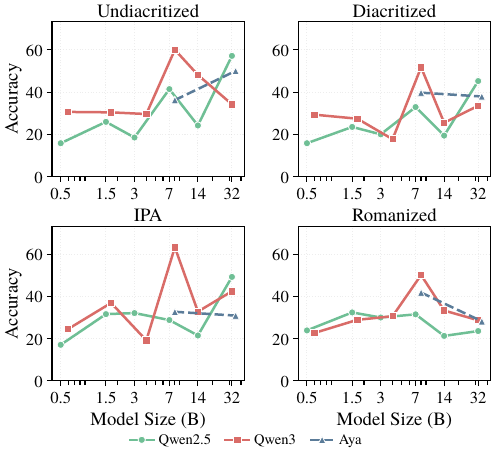}
    % \caption{Overall accuracy under different input representations for the Qwen2.5, Qwen3, and Aya models.}
    \caption{Effect of model scale on semantic disambiguation accuracy across different input representations for the Qwen2.5, Qwen3, and Aya-23 model families.}
    % \vspace{-1em}
    \label{fig:scaling_performance_sd}
\end{figure}

\paragraph{\colorbox{RQFourBg}{RQ4:} Impact of Model Size}
We compare multiple scales of Qwen2.5, Qwen3, and Aya-23, covering sub-billion to 35B-parameter models. Figure~\ref{fig:scaling_performance_ci} shows that scaling generally improves CI, but gains vary across model families and representations. For Qwen2.5 and Qwen3, performance improves substantially up to 14B under undiacritized and diacritized inputs, while scaling to 32B provides little additional benefit or slightly degrades performance. Gains are smaller under IPA and Romanization, where accuracy remains substantially lower. Aya shows modest gains with scale under undiacritized input, but little or no improvement under alternative representations.

SD exhibits a less consistent scaling pattern as shown in Figure~\ref{fig:scaling_performance_sd}. Qwen2.5 generally improves with scale, with large gains from 14B to 32B, whereas Qwen3 peaks at 8B across all representations and drops sharply at 14B. Aya again shows only modest changes across sizes. Scaling therefore benefits CI more under undiacritized inputs, while gains on SD are less predictable and depend strongly on model family and input representation.

\begin{figure}[t!]
    \centering
    \includegraphics[width=\linewidth]{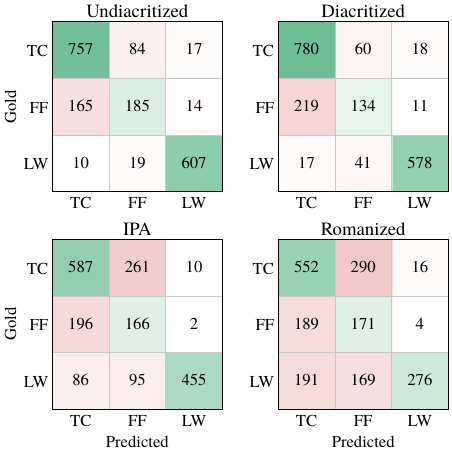}
%     \caption{Qwen3.6-Plus CI confusion matrices under four input representations.
% Rows are gold labels (TC/FF/LW) and columns are predictions.
% Cell values are raw counts.
% Color marks \textcolor{DeltaPos}{\textbf{correct}} vs.\ \textcolor{DeltaNeg}{\textbf{incorrect}}
% predictions; intensity scales with the absolute count.}

\caption{Qwen3.6-Plus confusion matrices for cognate identification across input representations. TC, FF, and LW denote true cognates, false friends, and loanwords.}
    \label{fig:qwen36_ci_cm}
\end{figure}

\subsection{Error Analysis}
\begin{figure}[t!]
    \centering
    \includegraphics[width=\linewidth]{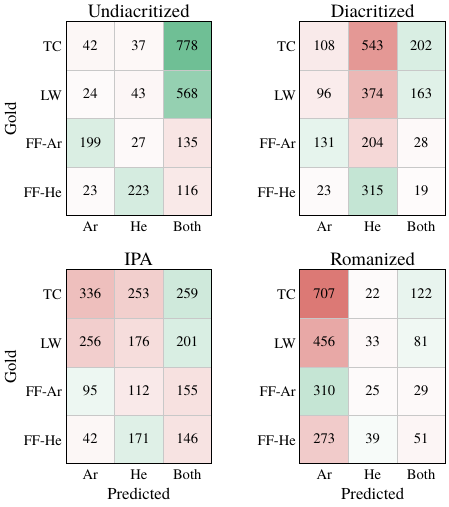}
    % \caption{Gemma-4-31B-IT SD confusion matrices under four input representations.
    % Each panel is a $4\times3$ matrix with gold rows TC, LW, FF-Ar, and FF-He
    % (predicted columns: Ar, He, Both).
    % Cell values are raw counts.
    % Color marks \textcolor{DeltaPos}{\textbf{correct}} vs.\ \textcolor{DeltaNeg}{\textbf{incorrect}}
    % predictions
    % (TC/LW~$\rightarrow$Both; FF-Ar~$\rightarrow$Ar; FF-He~$\rightarrow$He);
    % intensity scales with the absolute count.}
    \caption{Gemma-4-31B-IT confusion matrices for semantic disambiguation across input representations. TC/LW denote true cognates/loanwords; FF-Ar/FF-He denote whether only the Arabic/Hebrew sentence is correct.}
    \label{fig:gemma4_sd_cm}
\end{figure}

% \paragraph{Error Discussion}
\label{sec:error_discussion}
To better understand errors in the strongest LLMs, Figures~\ref{fig:qwen36_ci_cm} and~\ref{fig:gemma4_sd_cm} show confusion matrices for Qwen3.6-Plus, the best LLM on CI, and Gemma-4-31B-IT, the best on SD, across four input representations. In CI, errors are dominated by confusion between true cognates and false friends: true cognates are usually predicted as false friends, while false friends are overwhelmingly predicted as true cognates. Loanword errors differ: under undiacritized, diacritized, and IPA inputs, they are more often confused with false friends, whereas Romanization shifts them toward true cognates, further suggesting reliance on surface-form similarity.

% In CI, false friends remain the main challenge: under undiacritized input, 165 of 364 false friends are misclassified as true cognates, compared with 185 correctly identified. Diacritization increases confusion between false friends and true cognates. IPA and Romanization also reduce recognition of true cognates, while Romanization additionally leads to more loanword errors.

In SD, representation strongly affects which sentence the model favors. For true cognates and loanwords, where the correct answer is \textit{Both}, undiacritized input yields high accuracy; diacritization favors Hebrew, while IPA and Romanization shift errors toward Arabic. For false friends, undiacritized errors are mostly \textit{Both}. Diacritization improves Hebrew-only cases (223 to 315 correct) but shifts 204 Arabic-only cases to Hebrew-only. Under IPA, \textit{Both} again dominates, whereas Romanization favors Arabic, with 310 correct Arabic-only predictions and 273 Hebrew-only cases shifted to Arabic-only.
% A similar pattern appears in SD. Under undiacritized input, Gemma-4-31B-IT performs well when both sentences are valid, but often predicts \textit{Both} for false friends, effectively treating them as meaning-equivalent. Alternative representations also introduce strong answer biases, particularly toward Hebrew under diacritized input and Arabic under Romanization. Appendix~\ref{app:per-model-error-analysis} presents the same error analysis for each model.
Appendix~\ref{app:per-model-error-analysis} presents the same error analysis across CI and SD for each model.

% Overall, both tasks reveal the same failure mode: similar Arabic--Hebrew forms are often treated as semantically equivalent even when their meanings diverge.

% To characterize \emph{where} even the strongest systems fail, we inspect row-normalized confusion matrices for the best proprietary CI model (Qwen3.6-Plus; Figure~\ref{fig:qwen36_ci_cm}) and the best open-source SD model
% (Gemma-4-31B-IT; Figure~\ref{fig:gemma4_sd_cm}).
% In CI, loanwords and true cognates are largely recognized under native script, while false friends remain the main bottleneck and are most often confused with true cognates; diacritization further hurts FF, and IPA/romanization amplify TC--FF confusion and eventually degrade LW as well. In SD, Gemma-4-31B-IT is reliable when both sentences are valid (TC/LW), but false friends are much harder, with the dominant error being over-prediction of \textit{Both}; non-native representations introduce strong option biases. Overall, top models handle true cognate / loan word cases well in native script, but under-detect false friends: in CI, FF collapses into TC; in SD, FF collapses into \textit{Both}.

\section{Conclusion and Future Work}
We introduced SemCog Bench, a manually curated benchmark of 1,858 Arabic--Hebrew true cognates, false friends, and loanwords with word- and sentence-level annotations for cognate identification and semantic disambiguation. Across a diverse set of LLMs, we find strong reliance on surface-form similarity, with better performance on true cognates than false friends and wide variation on loanwords. Input representation, sentence-level context, and model scale all affect performance, but their effects vary across models and tasks. Original-script input is generally strongest, while context and scaling are more model-dependent.
% We introduced SemCog Bench, a manually benchmark for evaluating cross-lingual semantic reasoning in LLMs through Arabic--Hebrew true cognates, false friends, and loanwords. Across a diverse set of LLMs, we find that models rely heavily on surface-form similarity when reasoning about lexical relationships across languages. Although performance is generally strong on true cognates, models struggle to distinguish false friends and frequently over-generalize cognate relationships to modern loanwords. We also showed that neither sentence-level context nor increased model scale fully resolves these errors, and phonetic or Romanized representations further degrade performance.

In future work, we plan to extend SemCog Bench to additional language pairs and develop methods that better integrate orthographic, phonological, and contextual signals for cross-lingual lexical reasoning.

% SemCog Bench evaluates cross-lingual semantic reasoning in LLMs through Arabic--Hebrew true cognates, false friends, and loanwords. Across model families and input representations, our results show that current LLMs rely heavily on surface-form heuristics. Although models perform well on true cognates, they often misclassify false friends and over-project genetic relationships onto modern loanwords. Neither larger model scale nor sentence-level context fully resolves these form--meaning conflicts, and phonetic or Romanized inputs further disrupt performance. These findings highlight a key limitation in current multilingual LLMs and position SemCog Bench as a diagnostic benchmark for studying cross-lingual lexical interference.

\section*{Limitations}

SemCog Bench focuses on Arabic and Hebrew, so the findings may not generalize to language pairs with different typological, script, or contact patterns. Our evaluation uses controlled multiple-choice tasks, which enable systematic comparison but may not fully reflect open-ended settings such as generation, translation, or dialogue. The sentence contexts were initially generated with Gemini and subsequently reviewed and revised by human annotators; nevertheless, residual lexical or syntactic patterns from the generation process may remain. Finally, some input representations rely on automatic diacritization, Romanization, and IPA transcription tools, and proprietary-model results may vary with future model or API updates despite our use of fixed snapshots.

% This work has several limitations. First, SemCog Bench focuses on Arabic and Hebrew, and its findings may not directly generalize to language pairs with different typological, script, or contact patterns. Second, although the dataset covers true cognates, false friends, and loanwords, some borderline distinctions between inherited cognates and deeply integrated borrowings may still require linguistic judgment. We mitigate this ambiguity through expert annotation and additional verification. Third, our evaluation uses controlled multiple-choice formats, which facilitate systematic comparison but may not fully reflect performance in open-ended generation, translation, or dialogue settings. \textcolor{blue}{Fourth, the dataset includes 3,716 Gemini-generated sentences. Although human annotators independently reviewed all sentences for naturalness and revised them when necessary—retaining 3,549 (95.5\%) unchanged, making minor lexical edits to 151 (4.1\%), and substantially rewriting 16 (0.4\%)—this process cannot fully eliminate distribution bias such as subtle lexical or syntactic regularities inherited from Gemini that other LLMs might exploit.} Finally, some input representations rely on automatic diacritization, Romanization, and IPA transcription tools, and results for proprietary models may vary following future API updates despite our use of fixed model snapshots.

\section*{Ethics Statement}
This work uses linguistic resources and human-validated sentence examples to study Arabic--Hebrew lexical reasoning in LLMs. The dataset does not contain personal, private, or sensitive user information. Annotators have relevant linguistic expertise, and the annotation process includes independent verification of a subset of word-level labels. A key consideration is to avoid over-interpreting model behavior as reflecting properties of the languages or their speakers; our analysis is intended to diagnose limitations of models, not to evaluate linguistic communities. By identifying errors involving false friends and loanwords, this work aims to support more reliable multilingual systems for applications such as translation, language learning, and cross-lingual information access. 
% This work uses linguistic resources and human-validated sentence examples to study Arabic--Hebrew lexical reasoning in LLMs. The dataset does not contain personal, private, or sensitive user information. Annotators have relevant linguistic expertise, and disagreements are handled through verification and discussion. The main consideration is to avoid over-interpreting model behavior as reflecting properties of the languages or their speakers. Our analysis is intended to diagnose limitations of LLMs, not to evaluate linguistic communities. By identifying errors on false friends and loanwords, this work aims to support more reliable multilingual systems for tasks such as translation, language learning, and cross-lingual information access.
We used AI writing assistance within the scope of ``Assistance purely with the language of the paper'' described in the ACL Policy on Publication Ethics.

% We used AI writing assistance within the scope of ``Assistance purely with the language of the paper'' described in the ACL Policy on Publication Ethics.
% \subsection{Language-Specific Analysis}

\bibliography{custom,camel-bib-v3,anthology-1,anthology-2}

% \onecolumn
\newpage
\appendix

\section{Data Resources}
\label{appendix:data_resources}

We compiled candidate true cognates, false friends, and loanwords from a variety of dictionaries, linguistic databases, and academic publications.

\paragraph{True Cognates and False Friends}
Candidate true cognate and false friend pairs were collected from:
\begin{itemize}
    \item A Comparative Arabic--Hebrew Lexicon
    [\href{https://seveleu.com/pages/semitic-syntax-morpho/comparative-sem}{Source}].
    \item The Palisri Dictionary of the Palisra Project
    [\href{https://www.palisra.com/}{Source}].
\end{itemize}

\paragraph{Loanwords}
Candidate loanwords were collected from:
\begin{itemize}
    \item Hebrew loanword resources, including educational materials, Academy of the Hebrew Language documentation, and research on Modern Hebrew loanword phonology
    [\href{https://www.hebrewpod101.com/blog/2021/05/13/english-loanwords-in-hebrew/}{Source 1},
    \href{https://anglo-list.com/hebrew-words/}{Source 2},
    \href{https://hebrew-academy.org.il/%D7%9E%D7%99%D7%9C%D7%99%D7%9D-%D7%A9%D7%90%D7%95%D7%9C%D7%95%D7%AA-%D7%91%D7%A2%D7%91%D7%A8%D7%99%D7%AA/}{Source 3},
    \href{https://www.academia.edu/43719107/Loanword_phonology_in_Modern_Hebrew}{Source 4}].

    \item Arabic loanword resources, including studies of English loanwords, morphological adaptation, and lexical borrowing in Modern Standard Arabic
    [\href{https://www.meu.edu.jo/libraryTheses/24part2/English%20Loanwords%20in%20the%20Arab%20Newspapers.pdf}{Source 1},
    \href{https://www.researchgate.net/publication/340742540_THE_MORPHOLOGICAL_ADAPTATIONS_OF_ENGLISH_LOANWORDS_USED_IN_MODERN_STANDARD_ARABIC}{Source 2},
    \href{https://awej.org/images/AllIssues/Volume8/Volume8number3September/25.pdf}{Source 3},
    \href{https://www.researchgate.net/profile/Sami-Hamdi/publication/319434613_Lexical_Borrowing_in_Arabic_and_the_Role_of_Orthography/links/59e69a404585151e545ce2da/Lexical-Borrowing-in-Arabic-and-the-Role-of-Orthography.pdf}{Source 4}].
\end{itemize}

% To construct our lexicon of true cognates, false friends, and loanwords, we compiled resources from a diverse set of linguistic databases, dictionaries, and academic publications. 

% For the extraction of true cognates and false friends, we primarily relied on two comprehensive sources. First, we utilized a detailed Comparative Hebrew-Arabic lexicon [\href{https://seveleu.com/pages/semitic-syntax-morpho/comparative-sem}{TF1}]. Second, we consulted the dictionary of the Palisra project [\href{https://www.palisra.com/}{TF2}], an artistic and cultural initiative that features a constructed language described as an ``Arabic--Hebrew Esperanto.'' The Palisri dictionary includes extensively mapped Arabic and Hebrew vocabulary, providing a rich, aligned resource for our cognate annotation process.

% For the compilation of loanwords, we aggregated data from multiple language-specific repositories and linguistic studies. General lists of borrowed terms were sourced from collaborative databases for both Arabic [\href{https://en.wiktionary.org/wiki/Category:Arabic_borrowed_terms}{LW1}] and Hebrew [\href{https://en.wiktionary.org/wiki/Category:Hebrew_borrowed_terms}{LW2}]. For Hebrew specifically, we expanded our collection using pedagogical resources on English loanwords [\href{https://www.hebrewpod101.com/blog/2021/05/13/english-loanwords-in-hebrew/}{LW3}, \href{https://anglo-list.com/hebrew-words/}{LW4}], official documentation from the Academy of the Hebrew Language [\href{https://hebrew-academy.org.il/%D7%9E%D7%99%D7%9C%D7%99%D7%9D-%D7%A9%D7%90%D7%95%D7%9C%D7%95%D7%AA-%D7%91%D7%A2%D7%91%D7%A8%D7%99%D7%AA/}{LW5}], and academic analyses on Modern Hebrew loanword phonology [\href{https://www.academia.edu/43719107/Loanword_phonology_in_Modern_Hebrew}{LW6}]. For Arabic, our resources included comprehensive studies on English loanwords in Arab media [\href{https://www.meu.edu.jo/libraryTheses/24part2/English%20Loanwords%20in%20the%20Arab%20Newspapers.pdf}{LW7}], morphological adaptations of loanwords in Modern Standard Arabic [\href{https://www.researchgate.net/publication/340742540_THE_MORPHOLOGICAL_ADAPTATIONS_OF_ENGLISH_LOANWORDS_USED_IN_MODERN_STANDARD_ARABIC}{LW8}], and research investigating lexical borrowing and orthography [\href{https://awej.org/images/AllIssues/Volume8/Volume8number3September/25.pdf}{LW9}, \href{https://www.researchgate.net/profile/Sami-Hamdi/publication/319434613_Lexical_Borrowing_in_Arabic_and_the_Role_of_Orthography/links/59e69a404585151e545ce2da/Lexical-Borrowing-in-Arabic-and-the-Role-of-Orthography.pdf}{LW10}].

\section{LLM Snapshots and Licenses}
\label{apdx:model_snapshots}

\paragraph{Snapshots} To ensure the reproducibility and consistency of our experimental results, all evaluations involving proprietary models were executed using fixed, version-controlled API snapshots. Specifically, we used OpenAI's \texttt{gpt-5.4-2026-03-05} and \texttt{gpt-4o-2024-11-20}, DeepSeek's \texttt{deepseek-v4-pro}, and Alibaba Cloud's \texttt{qwen3.6-plus-2026-04-02}.

\paragraph{Licenses}
\begin{itemize}
    \item \textbf{Apache 2.0:}
    Qwen2.5-7B-Instruct~\cite{qwen2025qwen25technicalreport},
    Qwen3-8B~\cite{yang2025qwen3technicalreport},
    Gemma-4-31B-IT~\cite{gemmateam2024gemma2improvingopen},
    GPT-OSS-120B~\cite{openai2025gptoss},
    Jais-2-8B-Chat~\cite{anwar2026jais2familyarabiccentric},
    Jais-Adapted-7B-Chat~\cite{sengupta2023jais},
    and DictaLM-2.0-Instruct~\cite{shmidman2024adaptingllmshebrewunveiling}.

    \item \textbf{Llama Community License:}
    Llama-3.1-8B-Instruct and Llama-3.3-70B-Instruct~\cite{grattafiori2024llama3herdmodels}.

    \item \textbf{Gemma Terms of Use:}
    Gemma-2-9B-IT~\cite{gemmateam2024gemma2improvingopen}.

    \item \textbf{CC-BY-NC-4.0:}
    Aya-23-8B~\cite{aryabumi2024aya23openweight}.

    \item \textbf{GLM-4 License:}
    GLM-4-9B-Chat~\cite{glm2024chatglmfamilylargelanguage}.

    \item \textbf{MIT:}
    DeepSeek-v4-pro~\cite{deepseekai2025deepseekv3technicalreport}.

    \item \textbf{OpenAI Services Agreement:}
    GPT-5.4~\cite{gpt5} and GPT-4o~\cite{openai2024gpt4technicalreport}.

    \item \textbf{Alibaba Cloud Model Studio Terms:}
    Qwen3.6-Plus~\cite{yang2025qwen3technicalreport}.
\end{itemize}

\section{Compute and Hyperparameters}
\label{apdx:experiment_settings}

We evaluate the open-source models on two NVIDIA A40 GPUs, each with 48 GB of memory, and access the proprietary models through their official APIs. All experiments with open-source models are conducted using the LM Evaluation Harness framework\footnote{\url{https://github.com/EleutherAI/lm-evaluation-harness}}, which provides a consistent evaluation interface and supports reproducible comparisons.

For open-source model inference, we use the vLLM framework~\cite{kwon2023efficient} to improve computational efficiency and memory usage. To ensure reproducibility, we use deterministic decoding with temperature $t=0$ and limit the maximum output length to $max\_token=20$. For proprietary models, including GPT, DeepSeek, and Qwen, we use the same decoding configuration for $t$ and $max\_token$ whenever the corresponding API supports these parameters. We exclude the Gemini model family from the evaluation to avoid potential circular evaluation bias. To reduce position bias in the semantic disambiguation task, we randomly permute the order of answer choices and construct a fixed evaluation set with randomized option labels. All results reported in this work are computed on this fixed evaluation set.

\section{LLMs Prompts}
\label{appendix:evaluation_prompts}
\subsection{Cognate Identification}
% This section details the prompts designed for the cognate identification and semantic disambiguation tasks. These prompts incorporate various input formats, including undiacritized text, fully diacritized text, Uroman transliteration, and IPA transcription.

\begin{tcolorbox}[
    breakable,
    colback=white,
    colframe=red!60!black,
    title=\textbf{Cognate Identification (Undiacritized)},
    fontupper=\small,
]
\ttfamily
You are provided with an Arabic word and a Hebrew word. Your task is to classify the relationship between the words into one of the following categories:
- TRUE\_COGNATE
- FALSE\_FRIEND
- LOANWORD

Provide only one label as the response: TRUE\_COGNATE, FALSE\_FRIEND, or LOANWORD.

Definitions:

TRUE\_COGNATE: Words that share a common Semitic root and possess semantically overlapping meanings. At least one core meaning must be present in both languages.

FALSE\_FRIEND: Words that exhibit similar orthographic or phonological forms but convey completely different meanings, with no meaningful semantic overlap.

LOANWORD: Words that are borrowed from one language into another, resulting in similar forms without a shared Semitic root.

Arabic: \{arabic\_word\}

Hebrew: \{hebrew\_word\}

Answer:
\end{tcolorbox}

\begin{tcolorbox}[
  breakable,
  colback=white,
  colframe=red!60!black,
  title=\textbf{Cognate Identification (Diacritized)},
  fontupper=\small,
]
\ttfamily
You are provided with a fully diacritized Arabic word and a fully diacritized Hebrew word. Your task is to classify the relationship
between the words into one of the following categories:
- TRUE\_COGNATE
- FALSE\_FRIEND
- LOANWORD

Provide only one label as the response: TRUE\_COGNATE, FALSE\_FRIEND, or LOANWORD.

Definitions:

TRUE\_COGNATE: Words that share a common Semitic root and possess semantically overlapping meanings. At least one core meaning must be
present in both languages.

FALSE\_FRIEND: Words that exhibit similar diacritized orthographic or phonological forms (including vowel patterns and consonantal
skeletons) but convey completely different meanings, with no meaningful semantic overlap.

LOANWORD: Words that are borrowed from one language into another, resulting in similar diacritized forms without a shared Semitic root.

Arabic (diac): \{arabic\_word\_diac\}

Hebrew (diac): \{hebrew\_word\_diac\}

Answer:
\end{tcolorbox}

\begin{tcolorbox}[
  breakable,
  colback=white,
  colframe=red!60!black,
  title=\textbf{Cognate Identification (IPA)},
  fontupper=\small,
]
\ttfamily
You are provided with an IPA transcription of an Arabic word and a Hebrew word. Your task is to classify the relationship between the words
into one of the following categories:
- TRUE\_COGNATE
- FALSE\_FRIEND
- LOANWORD

Provide only one label as the response: TRUE\_COGNATE, FALSE\_FRIEND, or LOANWORD.

Definitions:

TRUE\_COGNATE: Words that share a common Semitic root and possess semantically overlapping meanings. At least one core meaning must be
present in both languages.

FALSE\_FRIEND: Words that exhibit similar phonetic forms in IPA transcription (reflecting similar pronunciation) but convey completely
different meanings, with no meaningful semantic overlap.

LOANWORD: Words that are borrowed from one language into another, resulting in similar IPA transcriptions without a shared Semitic root.

Arabic (IPA): \{arabic\_word\_ipa\}

Hebrew (IPA): \{hebrew\_word\_ipa\}

Answer:
\end{tcolorbox}

\begin{tcolorbox}[
  breakable,
  colback=white,
  colframe=red!60!black,
  title=\textbf{Cognate Identification (Romanized)},
  fontupper=\small,
]
\ttfamily
You are provided with a Universal Romanization of an Arabic word and a Hebrew word. Your task is to classify the relationship
between the words into one of the following categories:
- TRUE\_COGNATE
- FALSE\_FRIEND
- LOANWORD

Provide only one label as the response: TRUE\_COGNATE, FALSE\_FRIEND, or LOANWORD.

Definitions:

TRUE\_COGNATE: Words that share a common Semitic root and possess semantically overlapping meanings. At least one core meaning must be
present in both languages.

FALSE\_FRIEND: Words that exhibit similar romanized forms (reflecting similar phonological or orthographic structure) but convey completely
different meanings, with no meaningful semantic overlap.

LOANWORD: Words that are borrowed from one language into another, resulting in similar romanized forms without a shared Semitic root.

Arabic (uroman): \{arabic\_word\_uroman\}

Hebrew (uroman): \{hebrew\_word\_uroman\}

Answer:
\end{tcolorbox}

\begin{tcolorbox}[
    breakable,
  colback=white,
  colframe=red!60!black,
    title=\textbf{Cognate Identification (Undiacritized with Context)},
    fontupper=\small,
]
\ttfamily
You are given an Arabic sentence and a Hebrew sentence, along with a specific target Arabic word and a target Hebrew word extracted from these sentences.

Your task is to determine the relationship between the target Arabic word and the target Hebrew word, taking into account their context within the sentences.

Determine whether they are:
- TRUE\_COGNATE
- FALSE\_FRIEND
- LOANWORD

Answer with only one label: TRUE\_COGNATE, FALSE\_FRIEND, or LOANWORD.

Definitions:

TRUE\_COGNATE:
Words that share a common Semitic root AND have semantically overlapping meanings in the given context.
At least one core meaning must be present in both languages.

FALSE\_FRIEND:
Words that have similar orthographic or phonological forms BUT have completely different meanings in the given context, with no meaningful semantic overlap.

LOANWORD:
Words that are borrowed from one language into another, resulting in similar forms without a shared Semitic root.

Arabic Sentence: \{arabic\_sentence\}
Target Arabic Word: \{arabic\_word\}

Hebrew Sentence: \{hebrew\_sentence\}
Target Hebrew Word: \{hebrew\_word\}

Answer:
\end{tcolorbox}

\subsection{Semantic Disambiguation}
\begin{tcolorbox}[
  breakable,
  colback=white,
  colframe=red!60!black,
  title=\textbf{Semantic Disambiguation (Undiacritized)},
  fontupper=\small,
]
\ttfamily
You are provided with two sentences: one written in the Arabic script and one in the Hebrew script. Which sentence is semantically
appropriate?

A. \{sent\_a\}

B. \{sent\_b\}

C. Both sentences are appropriate.

Provide only the corresponding letter (A, B, or C) as the response.

Answer:
\end{tcolorbox}

\begin{tcolorbox}[
  breakable,
  colback=white,
  colframe=red!60!black,
  title=\textbf{Semantic Disambiguation (Diacritized)},
  fontupper=\small,
]
\ttfamily
You are provided with two fully diacritized sentences: one in Arabic script and one in Hebrew script. Which sentence is semantically
appropriate?

A. \{sent\_a\}

B. \{sent\_b\}

C. Both sentences are appropriate.

Provide only the corresponding letter (A, B, or C) as the response.

Answer:
\end{tcolorbox}

\begin{tcolorbox}[
  breakable,
  colback=white,
  colframe=red!60!black,
  title=\textbf{Semantic Disambiguation (IPA)},
  fontupper=\small,
]
\ttfamily
You are provided with two sentences in IPA transcription: one from Arabic and one from Hebrew. Which sentence is semantically appropriate?

A. \{sent\_a\}

B. \{sent\_b\}

C. Both sentences are appropriate.

Provide only the corresponding letter (A, B, or C) as the response.

Answer:
\end{tcolorbox}

\begin{tcolorbox}[
  breakable,
  colback=white,
  colframe=red!60!black,
  title=\textbf{Semantic Disambiguation (Romanized)},
  fontupper=\small,
]
\ttfamily
You are provided with two sentences in transliteration by uroman: one from Arabic and one from Hebrew. Which sentence is semantically
appropriate?

A. \{sent\_a\}

B. \{sent\_b\}

C. Both sentences are appropriate.

Provide only the corresponding letter (A, B, or C) as the response.

Answer:
\end{tcolorbox}

\subsection{Sentence Generation}
\label{apdx:sentence_generation_prompt}
\begin{tcolorbox}[
    breakable,
  colback=white,
  colframe=red!60!black,
    title=\textbf{Sentence Generation},
    fontupper=\small,
]
\ttfamily
You are provided with an Arabic word, a Hebrew word, their corresponding meanings, and a cognate type. Your task is to generate one natural Arabic sentence and one natural Hebrew sentence, each containing the corresponding target word.

Inputs:

Arabic word: \{arabic\_word\}

Hebrew word: \{hebrew\_word\}

Arabic meaning: \{arabic\_meaning\}

Hebrew meaning: \{hebrew\_meaning\}

Cognate type: \{cognate\_type\}

Output format:

Return only valid JSON with the following fields: arabic\_word, hebrew\_word, arabic\_sentence, hebrew\_sentence, arabic\_translation\_en, and hebrew\_translation\_en.

Sentence requirements:

Each sentence must contain 8--20 words.

Each sentence must be natural, fluent, and representative of native-speaker usage.

The target word must be clearly used in its given meaning. Inflection is allowed.

Cognate constraints:

If cognate\_type is true\_cognate, both target words should be used with the same meaning, but the Arabic and Hebrew sentences should describe different situations.

If cognate\_type is false\_friend, the Arabic sentence must reflect the Arabic meaning, and the Hebrew sentence must reflect the Hebrew meaning. The difference between the two meanings must be clear from context.

Answer:
\end{tcolorbox}

\clearpage
\onecolumn
\section{Per-Model Error Analysis}
\label{app:per-model-error-analysis}
\subsection{Cognate Identification}

\begin{figure*}[ht!]
    \centering
    \includegraphics[width=\linewidth]{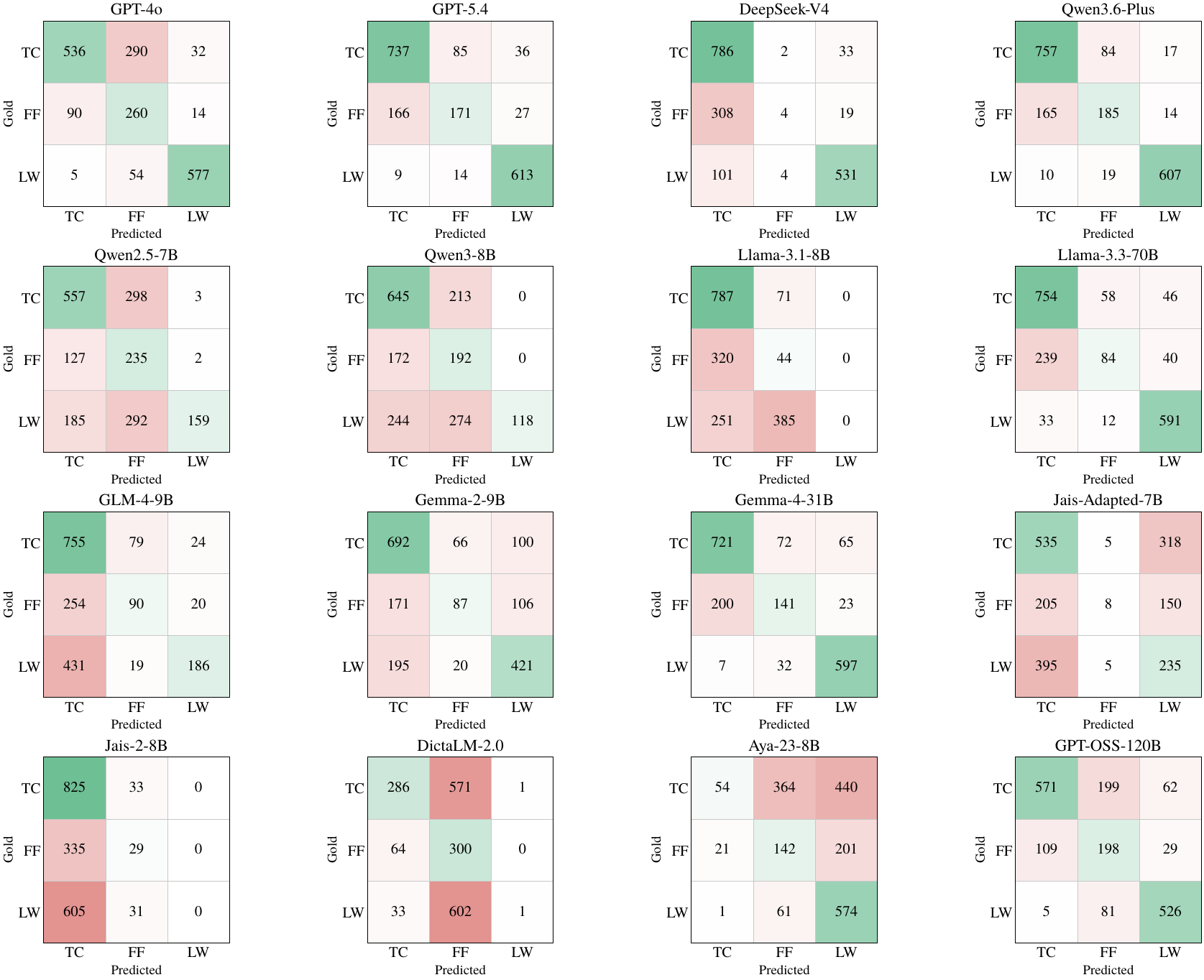}
    % \caption{Per-model CI confusion matrices under the undiacritized representation
    % (3-class setting).
    % Each panel is a $3\times3$ matrix for one model:
    % rows are gold labels (TC/FF/LW) and columns are predictions.
    % Cell values are raw counts.
    % Color marks \textcolor{DeltaPos}{correct} (diagonal) vs.\ \textcolor{DeltaNeg}{incorrect}
    % predictions; intensity scales with the absolute count.}
    \caption{Per-model confusion matrices for cognate identification using original undiacritized input. TC, FF, and LW denote true cognates, false friends, and loanwords.}
    \label{fig:ci_confusion_appendix_undiac}
\end{figure*}

\begin{figure*}[th!]
    \centering
    \includegraphics[width=\linewidth]{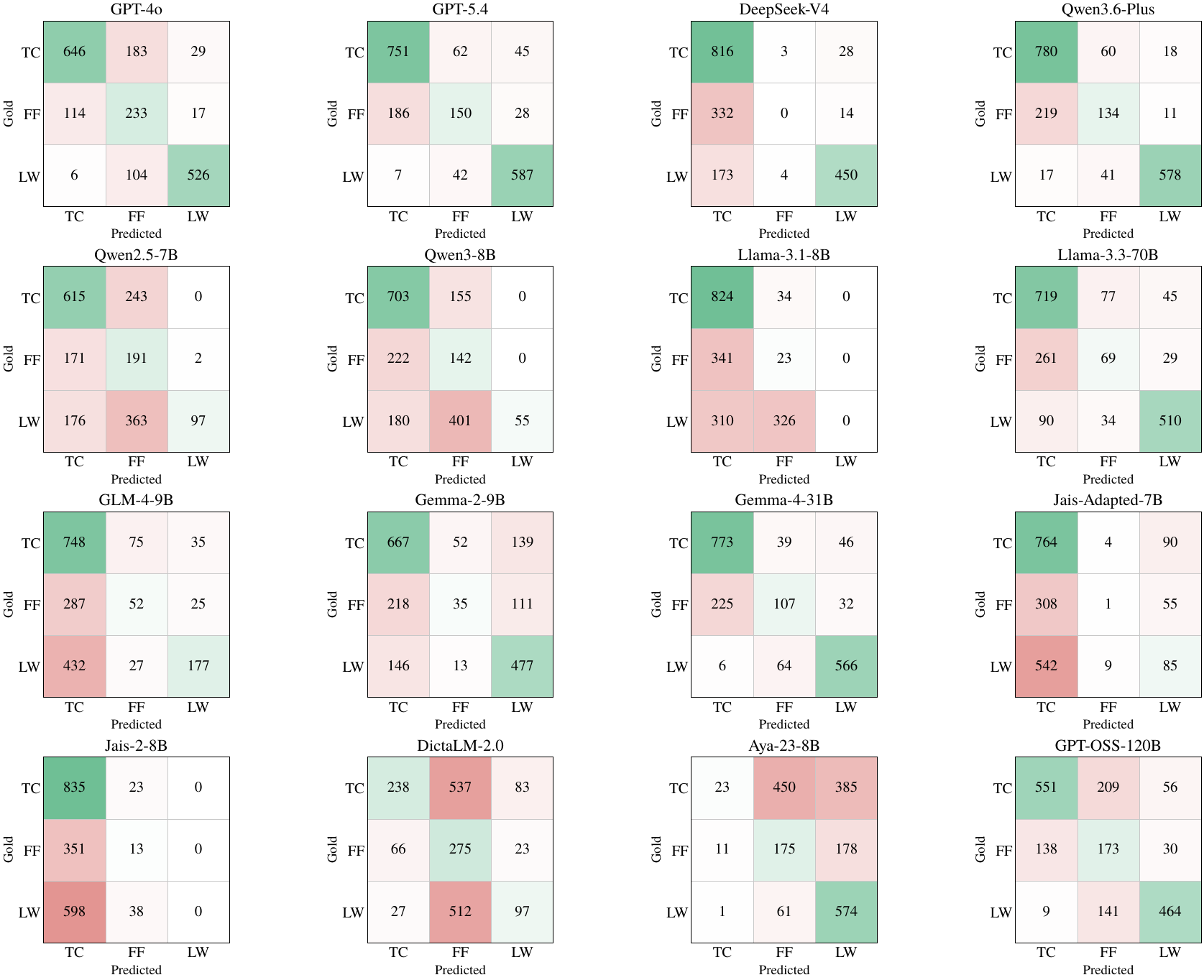}
    \caption{Per-model confusion matrices for cognate identification using diacritized input. TC, FF, and LW denote true cognates, false friends, and loanwords.}
    \label{fig:ci_confusion_appendix_diac}
\end{figure*}

\begin{figure*}[t]
    \centering
    \includegraphics[width=\linewidth]{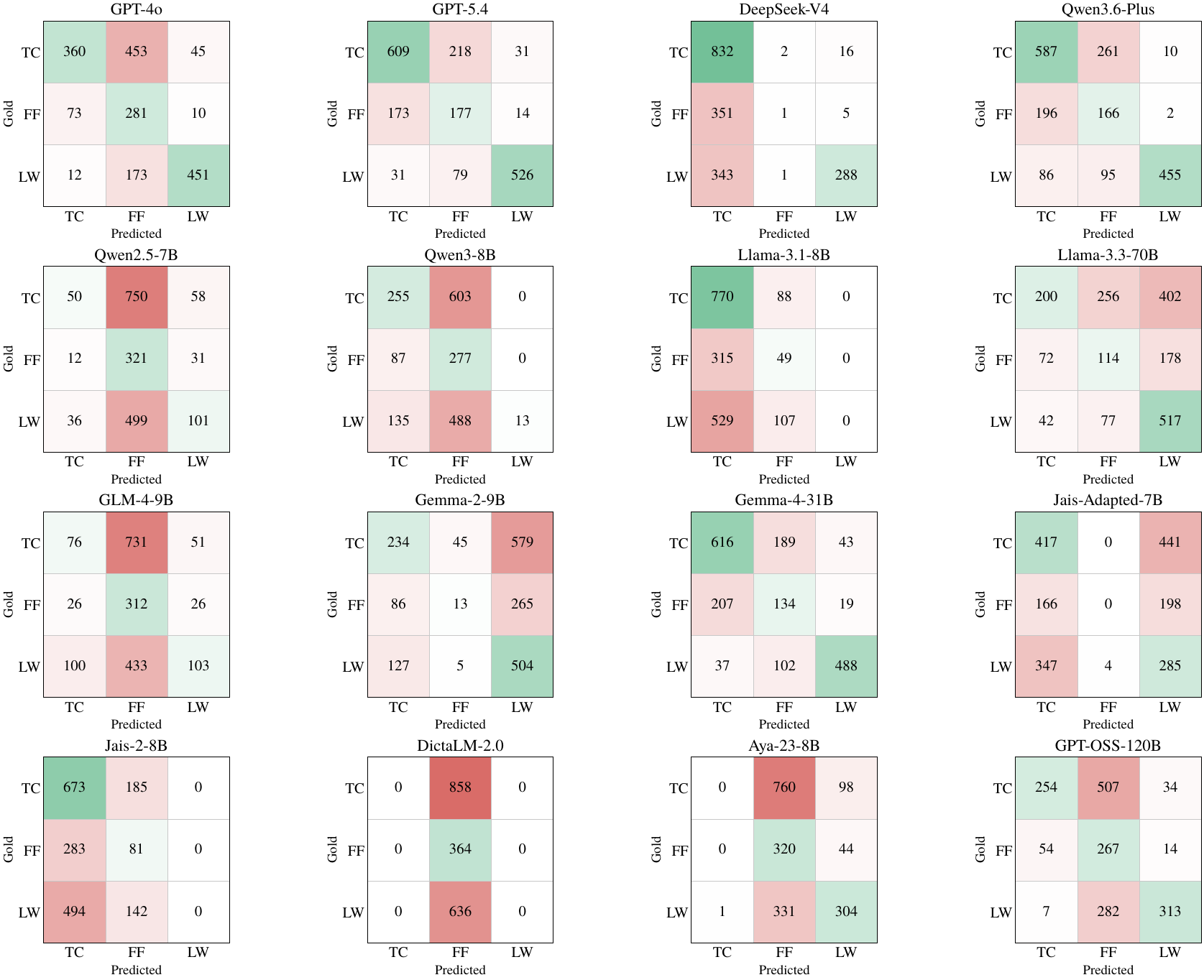}
    \caption{Per-model confusion matrices for cognate identification using IPA input. TC, FF, and LW denote true cognates, false friends, and loanwords.}
    \label{fig:ci_confusion_appendix_ipa}
\end{figure*}

\begin{figure*}[t]
    \centering
    \includegraphics[width=\linewidth]{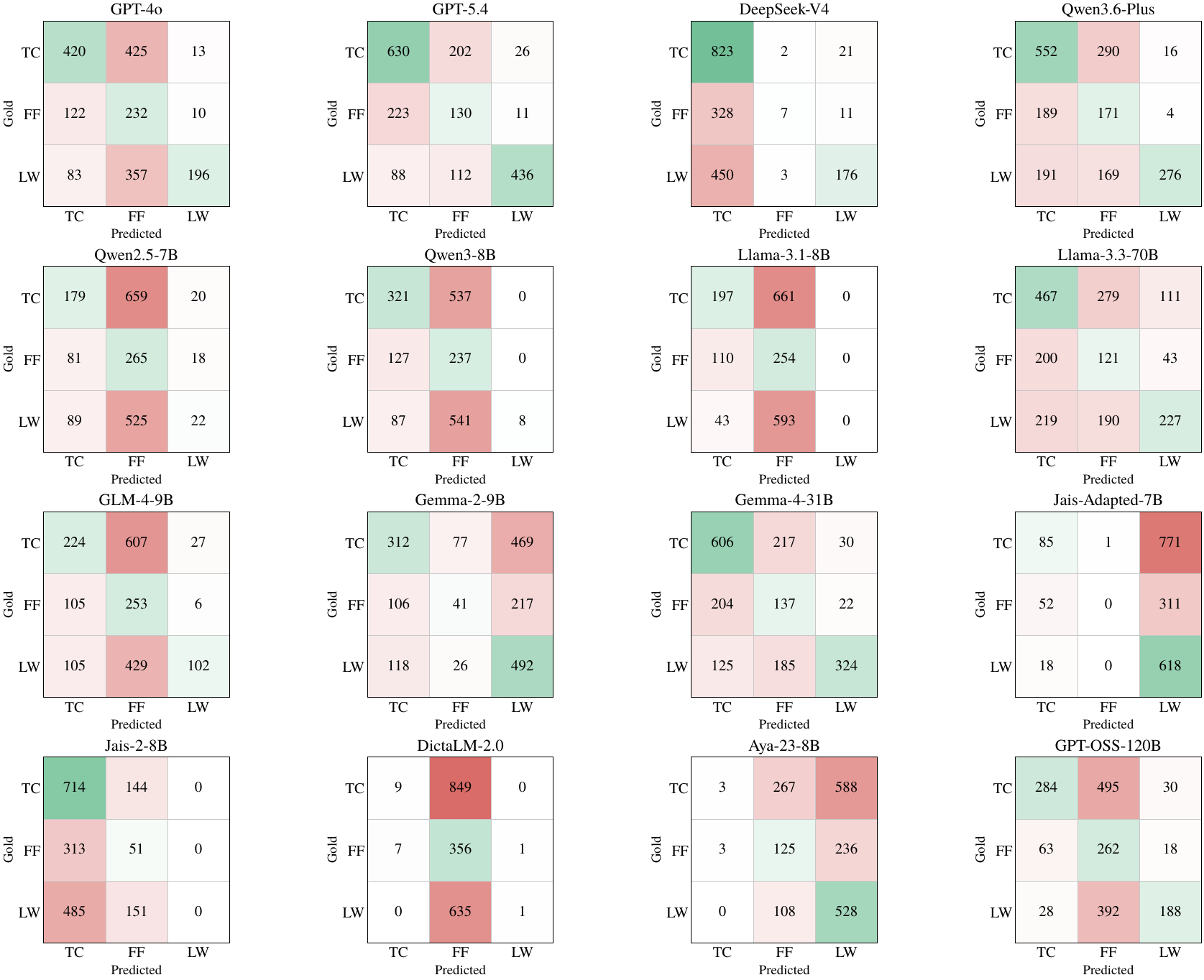}
    \caption{Per-model confusion matrices for cognate identification using Romanized input. TC, FF, and LW denote true cognates, false friends, and loanwords.}
    \label{fig:ci_confusion_appendix_uroman}
\end{figure*}

% \paragraph{Semantic Disambiguation.}
% Because SD items are constructed differently across lexical types, we report
% matrices separately for TC, FF, and LW
% (Figures~\ref{fig:sd_confusion_tc}--\ref{fig:sd_confusion_lw}).
% Here labels denote semantic judgments rather than MCQ slots:
% \textit{Ar} = only the Arabic sentence is appropriate,
% \textit{He} = only the Hebrew sentence is appropriate, and
% \textit{Both} = both sentences are appropriate.
% For TC and LW, the gold label is always \textit{Both}; for FF, it is either
% \textit{Ar} or \textit{He}.
% Across models, a recurring failure mode is over-predicting \textit{Both}
% when only one sentence is correct (FF), or selecting a single-sentence option
% when both are correct (TC/LW).

\clearpage
\subsection{Semantic Disambiguation}
\begin{figure*}[th!]
    \centering
    \includegraphics[width=\linewidth]{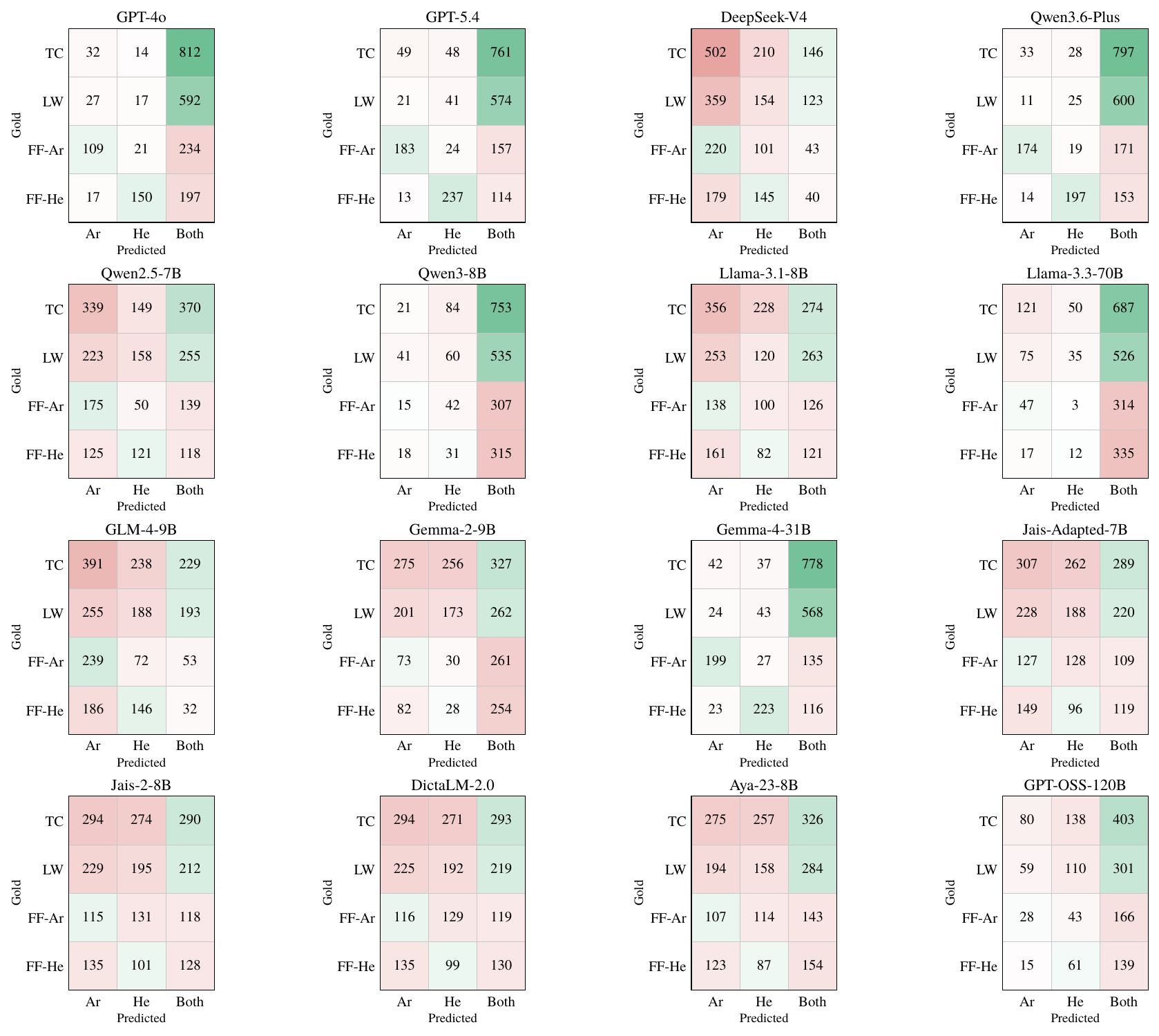}
    \caption{Per-model confusion matrices for semantic disambiguation with undiacritized input. TC/LW denote true cognates/loanwords; FF-Ar and FF-He indicate whether only the Arabic or Hebrew sentence is correct.}
%     \caption{Per-model SD confusion matrices under the undiacritized representation.
% Each panel is a $4\times3$ matrix: rows are gold labels and columns are predictions among
% \textit{Ar} (only the Arabic sentence is appropriate),
% \textit{He} (only the Hebrew sentence is appropriate), and
% \textit{Both} (both sentences are appropriate).
% The first two rows correspond to the true-cognate (TC) and loanword (LW)
% subtasks, where the gold label is always \textit{Both};
% the last two rows correspond to the false-friend (FF) subtask, where the
% gold label is either \textit{Ar} (\textit{FF-Ar}) or \textit{He} (\textit{FF-He}).
% Cell values are raw counts.
% Color marks \textcolor{DeltaPos}{correct} vs.\ \textcolor{DeltaNeg}{incorrect}
% predictions
% (TC/LW~$\rightarrow$\textit{Both}; FF-Ar~$\rightarrow$\textit{Ar}; FF-He~$\rightarrow$\textit{He});
% intensity scales with the absolute count.}
    \label{fig:sd_confusion_combined_undiac}
\end{figure*}
% \vspace{-50pt}
\begin{figure*}[th!]
    \centering
    \includegraphics[width=\linewidth]{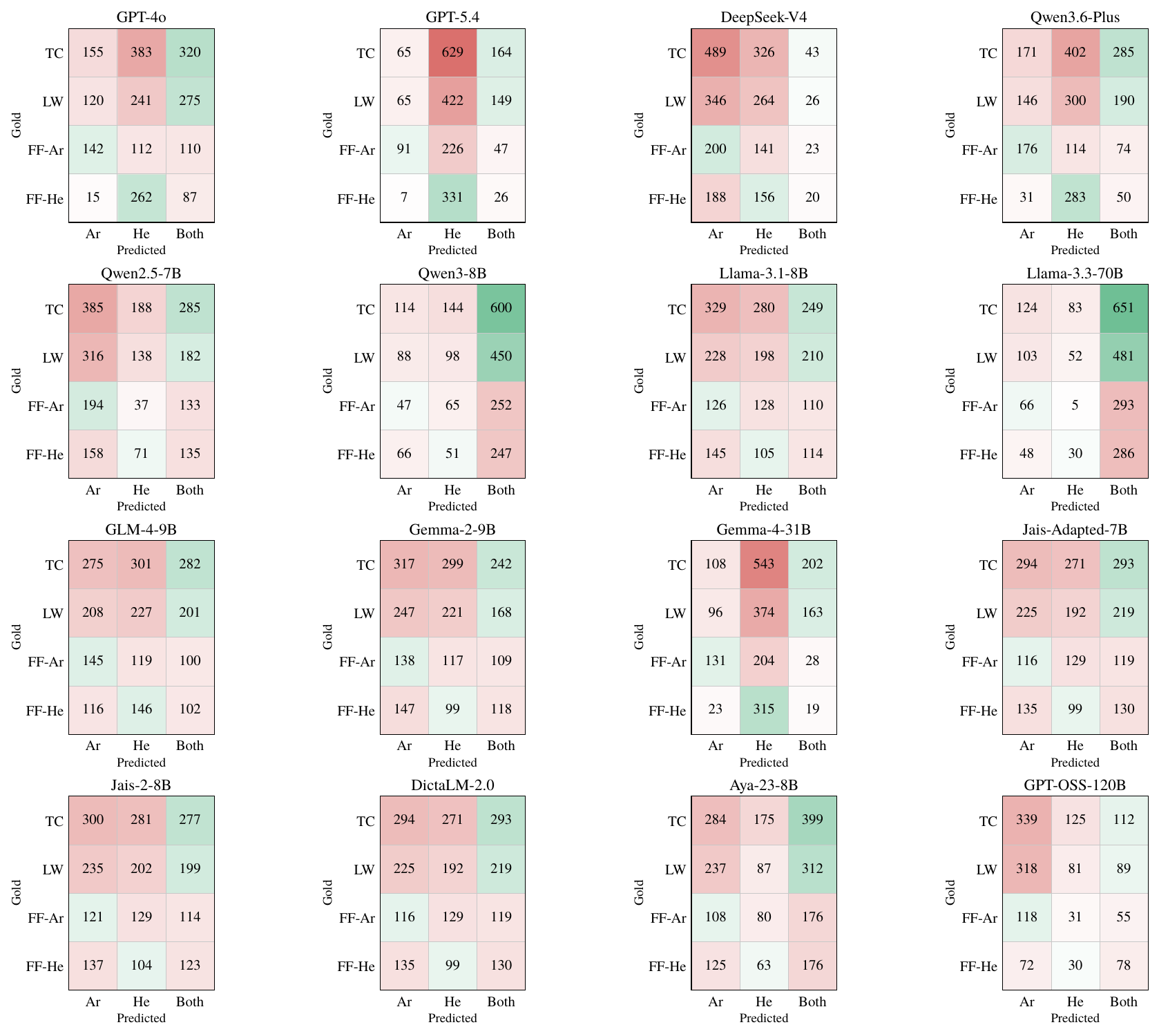}
    \caption{Per-model confusion matrices for semantic disambiguation with diacritized input. TC/LW denote true cognates/loanwords; FF-Ar and FF-He indicate whether only the Arabic or Hebrew sentence is correct.}
    % \caption{Per-model SD confusion matrices under the diacritized representation
    % (same $4\times3$ layout as Figure~\ref{fig:sd_confusion_combined_undiac}).}
    \label{fig:sd_confusion_combined_diac}
\end{figure*}

\begin{figure*}[t]
    \centering
    \includegraphics[width=\linewidth]{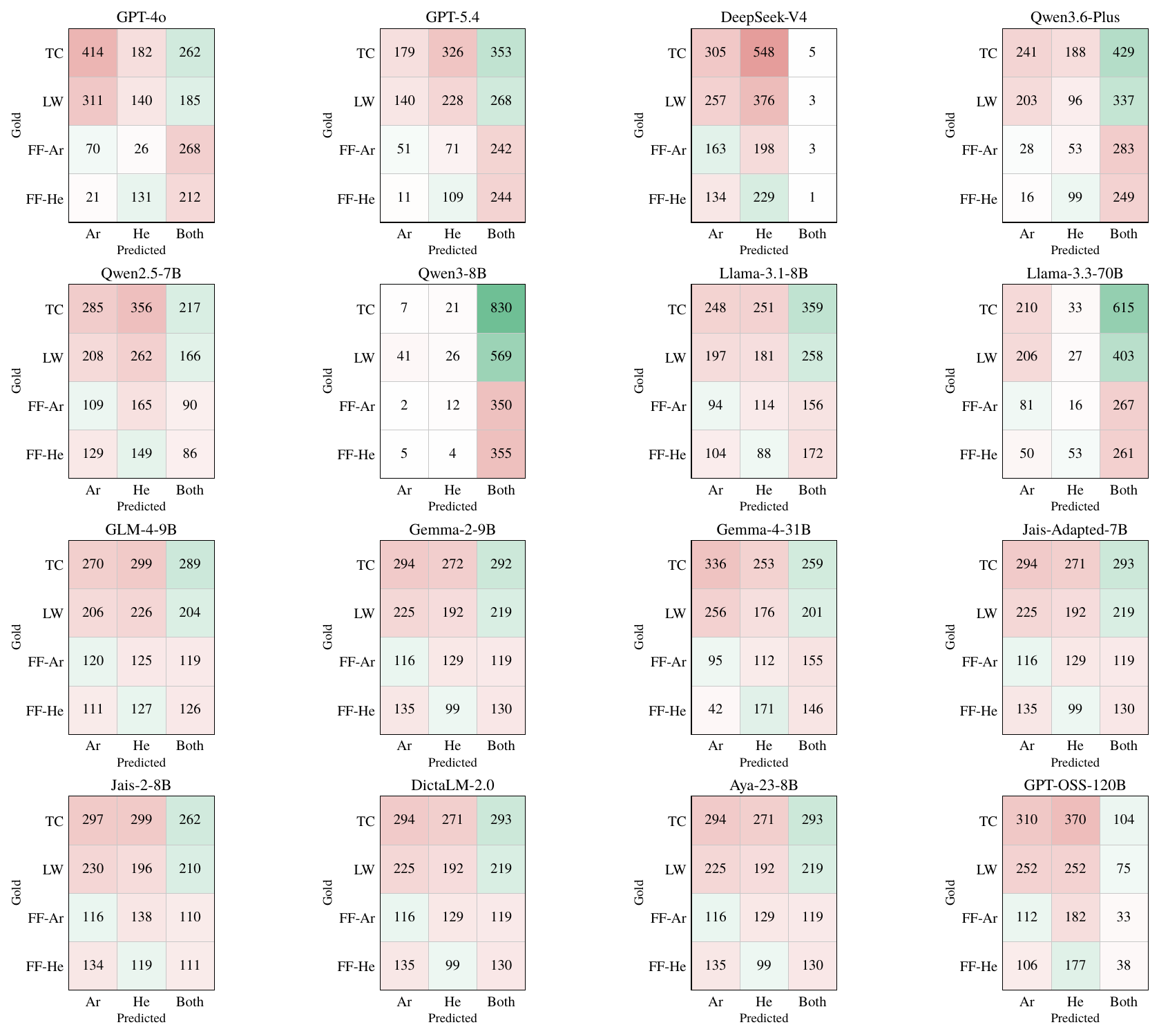}
    % \caption{Per-model SD confusion matrices under the IPA representation
    % (same $4\times3$ layout as Figure~\ref{fig:sd_confusion_combined_undiac}).}
    \caption{Per-model confusion matrices for semantic disambiguation with IPA input. TC/LW denote true cognates/loanwords; FF-Ar and FF-He indicate whether only the Arabic or Hebrew sentence is correct.}
    \label{fig:sd_confusion_combined_ipa}
\end{figure*}

\begin{figure*}[t]
    \centering
    \includegraphics[width=\linewidth]{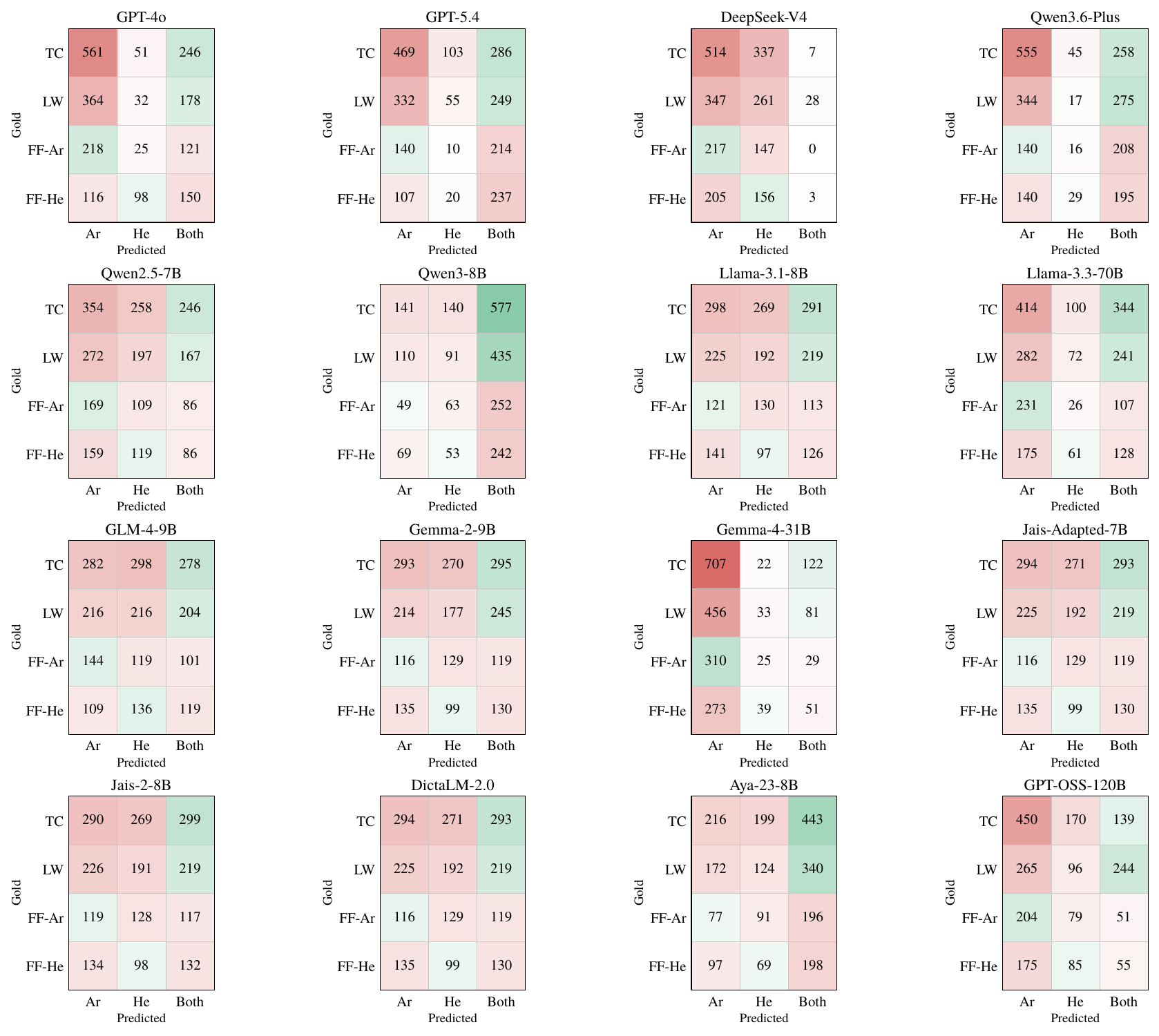}
    % \caption{Per-model SD confusion matrices under the Uroman representation
    % (same $4\times3$ layout as Figure~\ref{fig:sd_confusion_combined_undiac}).}
    \caption{Per-model confusion matrices for semantic disambiguation with Romanized input. TC/LW denote true cognates/loanwords; FF-Ar and FF-He indicate whether only the Arabic or Hebrew sentence is correct.}
    \label{fig:sd_confusion_combined_uroman}
\end{figure*}

% \section{AI Assistant Use}

% In producing this work, we used Grammarly for minor grammar and spelling corrections. We also used ChatGPT and Claude Code to assist with LaTeX table and figure formatting, code troubleshooting, and resolving issues related to Matplotlib visualizations. All suggestions generated by these tools were carefully reviewed and verified by the authors before being incorporated into the paper.

\clearpage
\onecolumn
\section{Annotation Guidelines}
\label{app:annotation-guidlines}
\setlength{\fboxrule}{0.2pt} % border thickness
\setlength{\fboxsep}{0pt}    % padding between border and image

\begin{center}
    \fbox{\includegraphics[width=\textwidth]{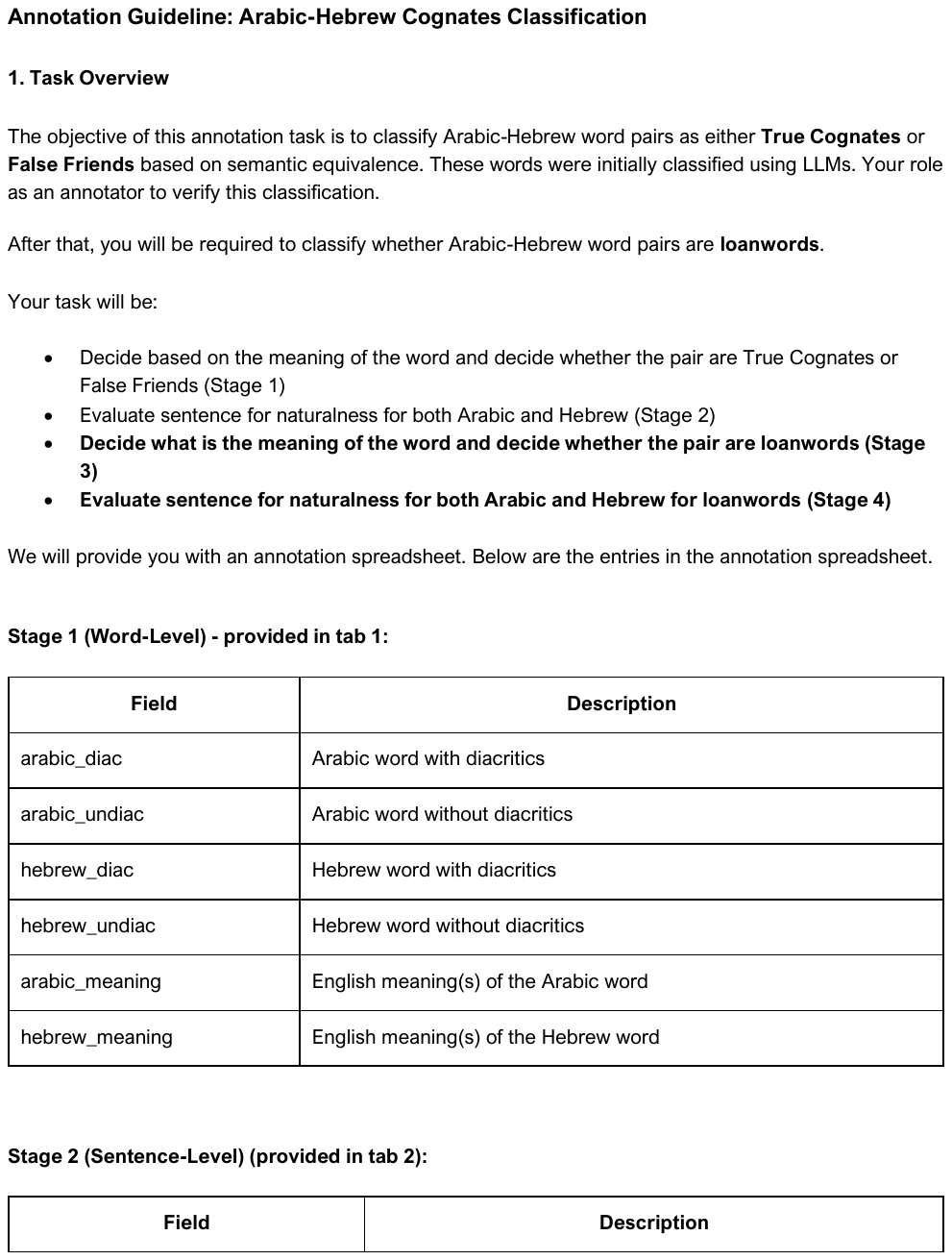}}

    \fbox{\includegraphics[width=\textwidth]{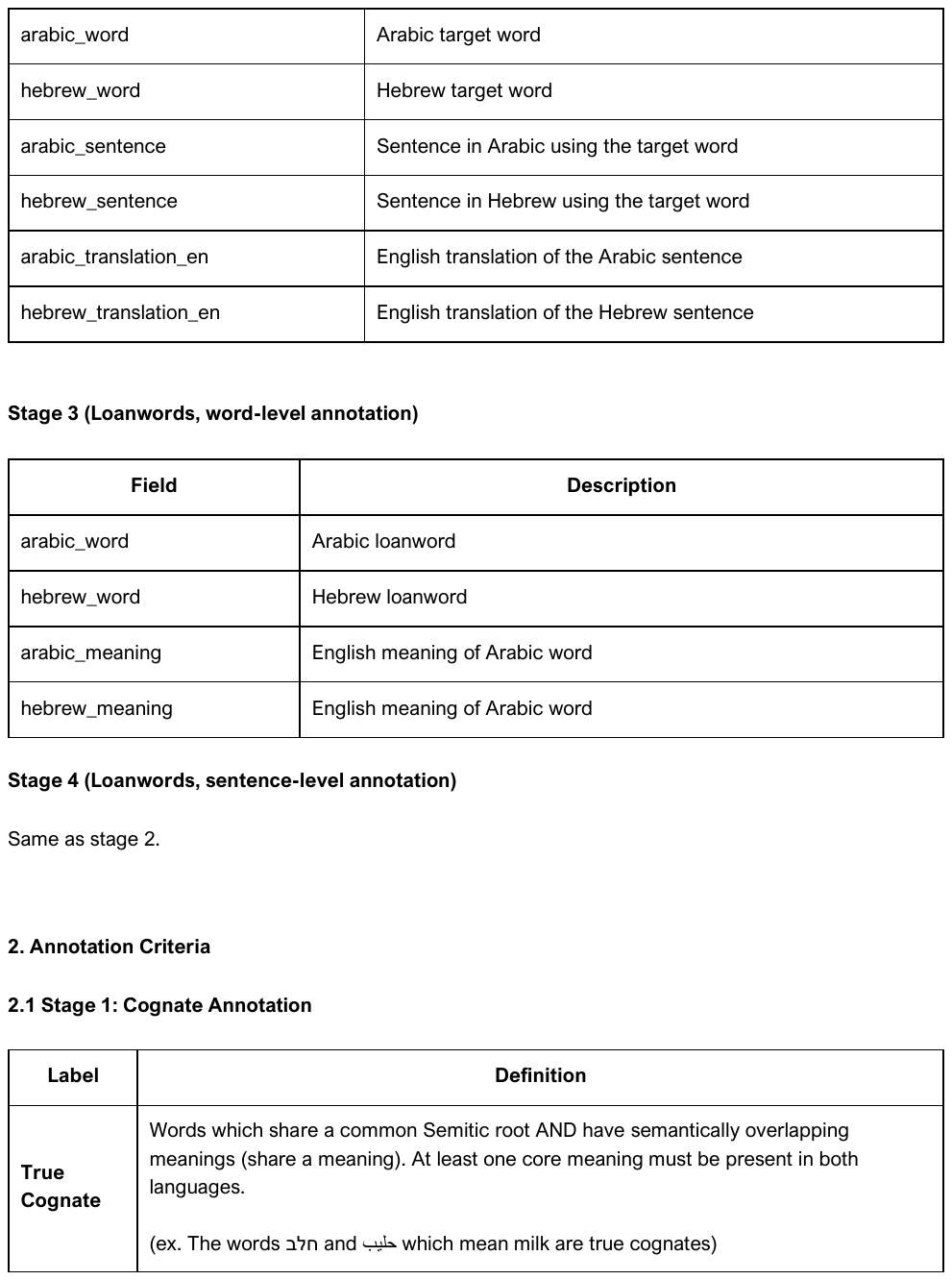}}

    \fbox{\includegraphics[width=\textwidth]{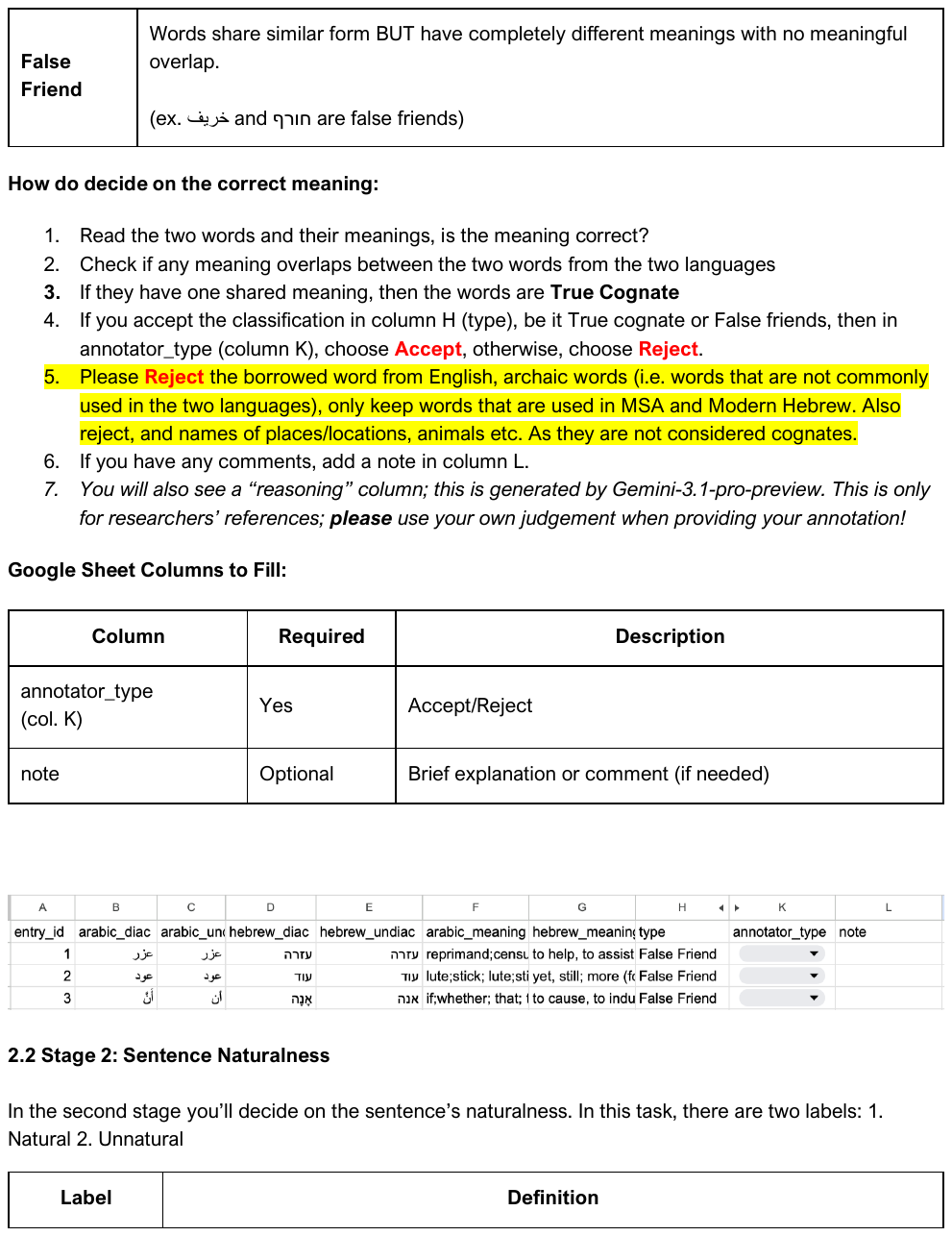}}

    \fbox{\includegraphics[width=\textwidth]{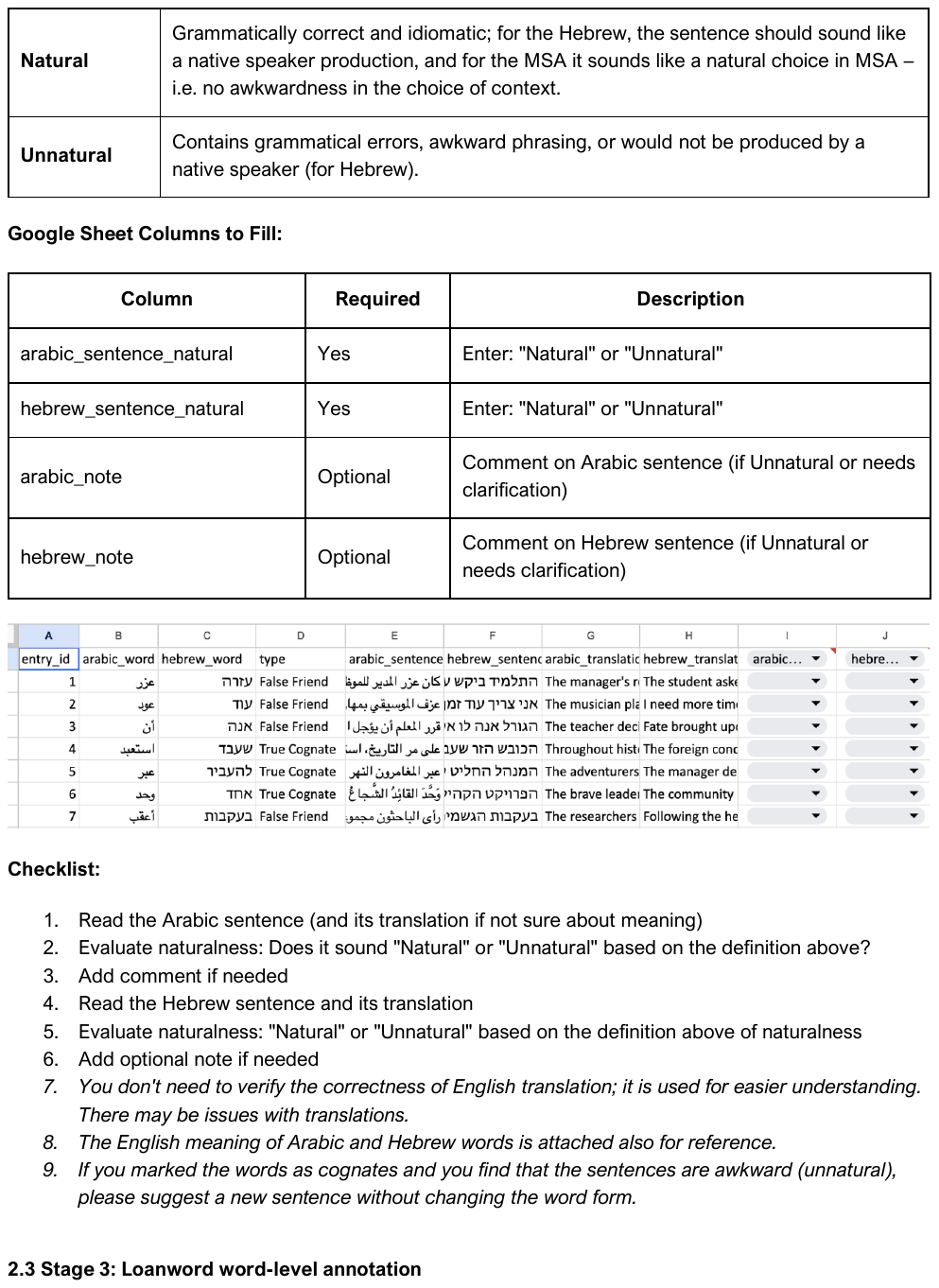}}

    \fbox{\includegraphics[width=\textwidth]{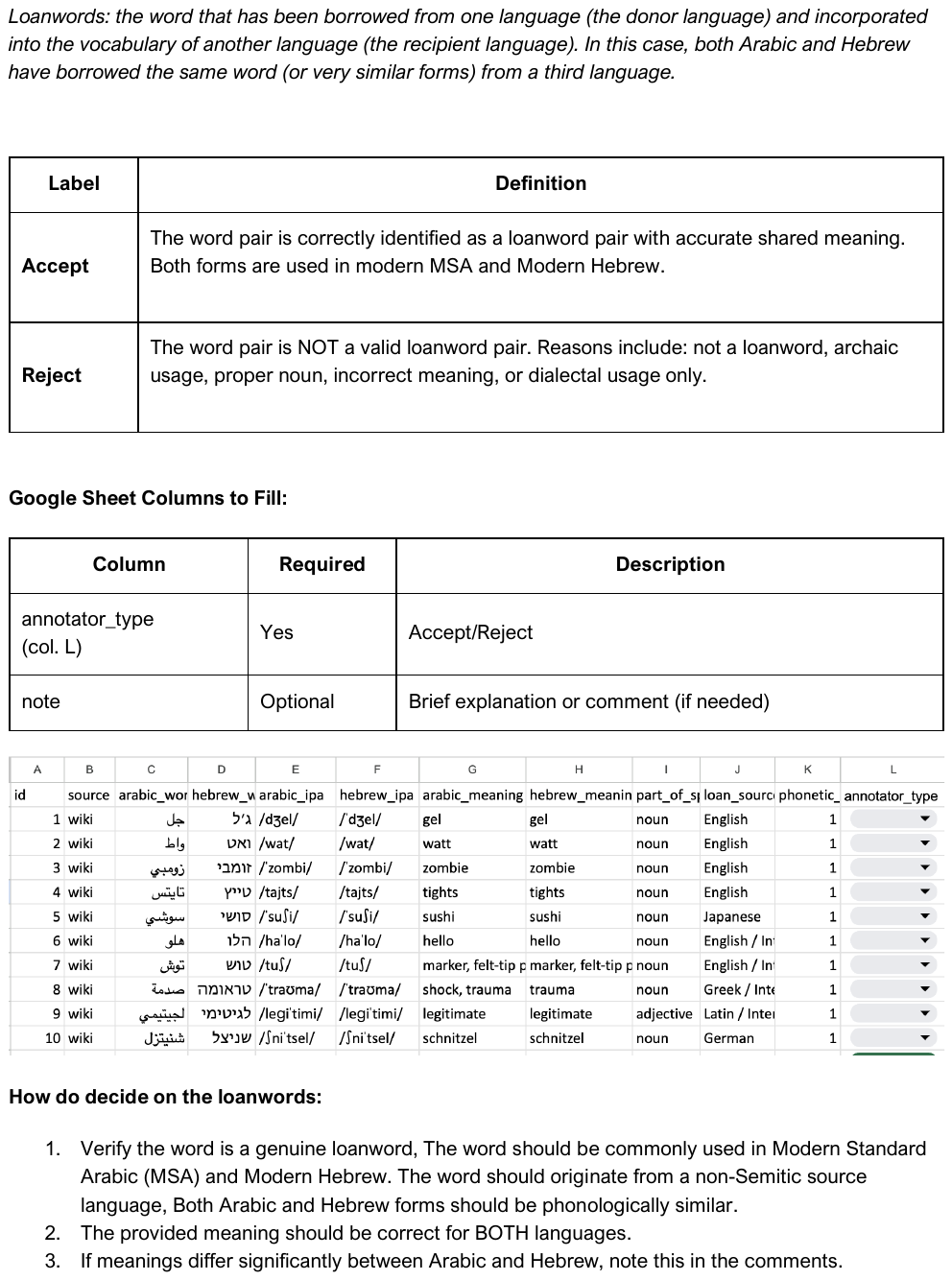}}

    \fbox{\includegraphics[width=\textwidth]{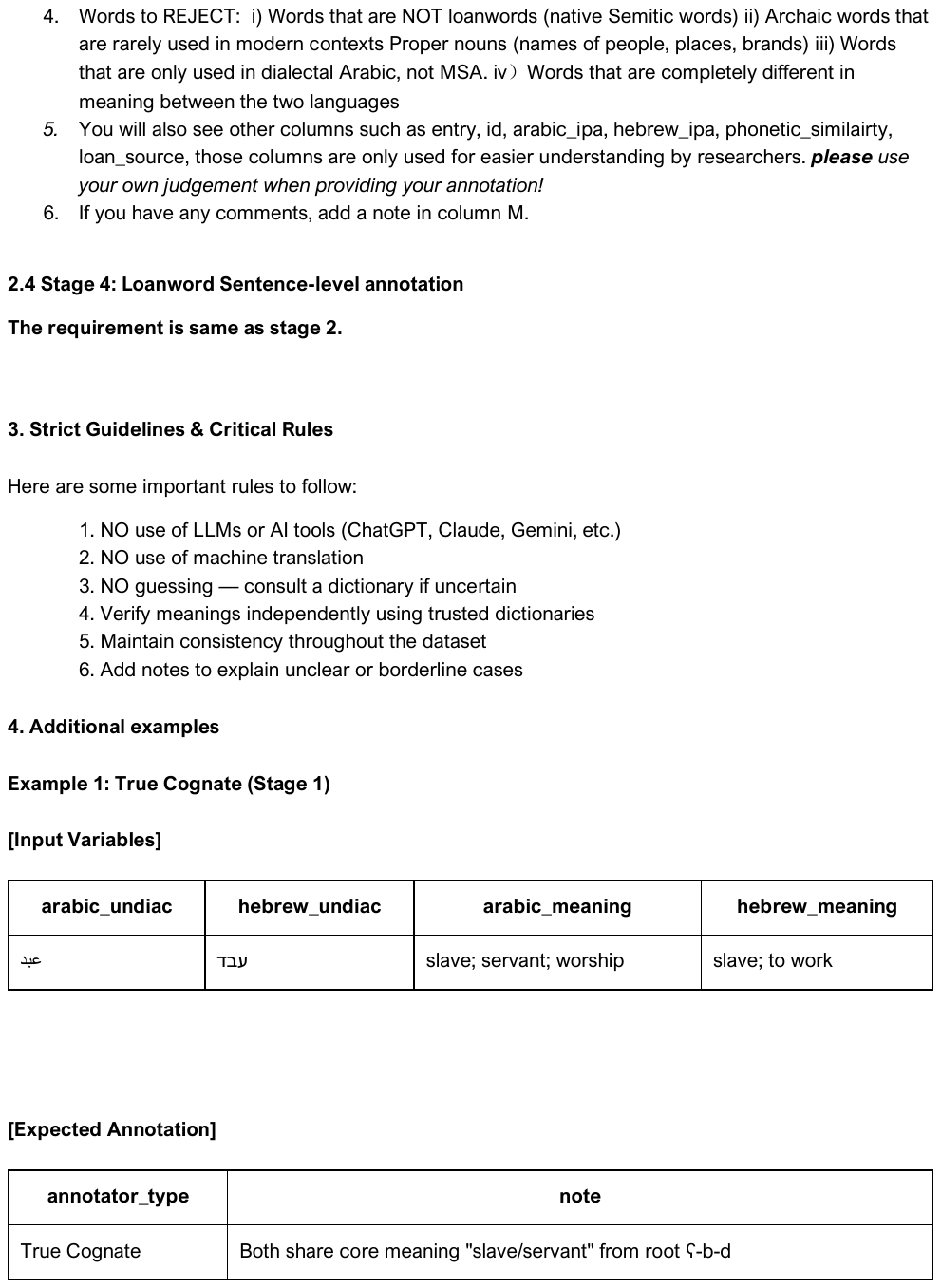}}

    \fbox{\includegraphics[width=\textwidth]{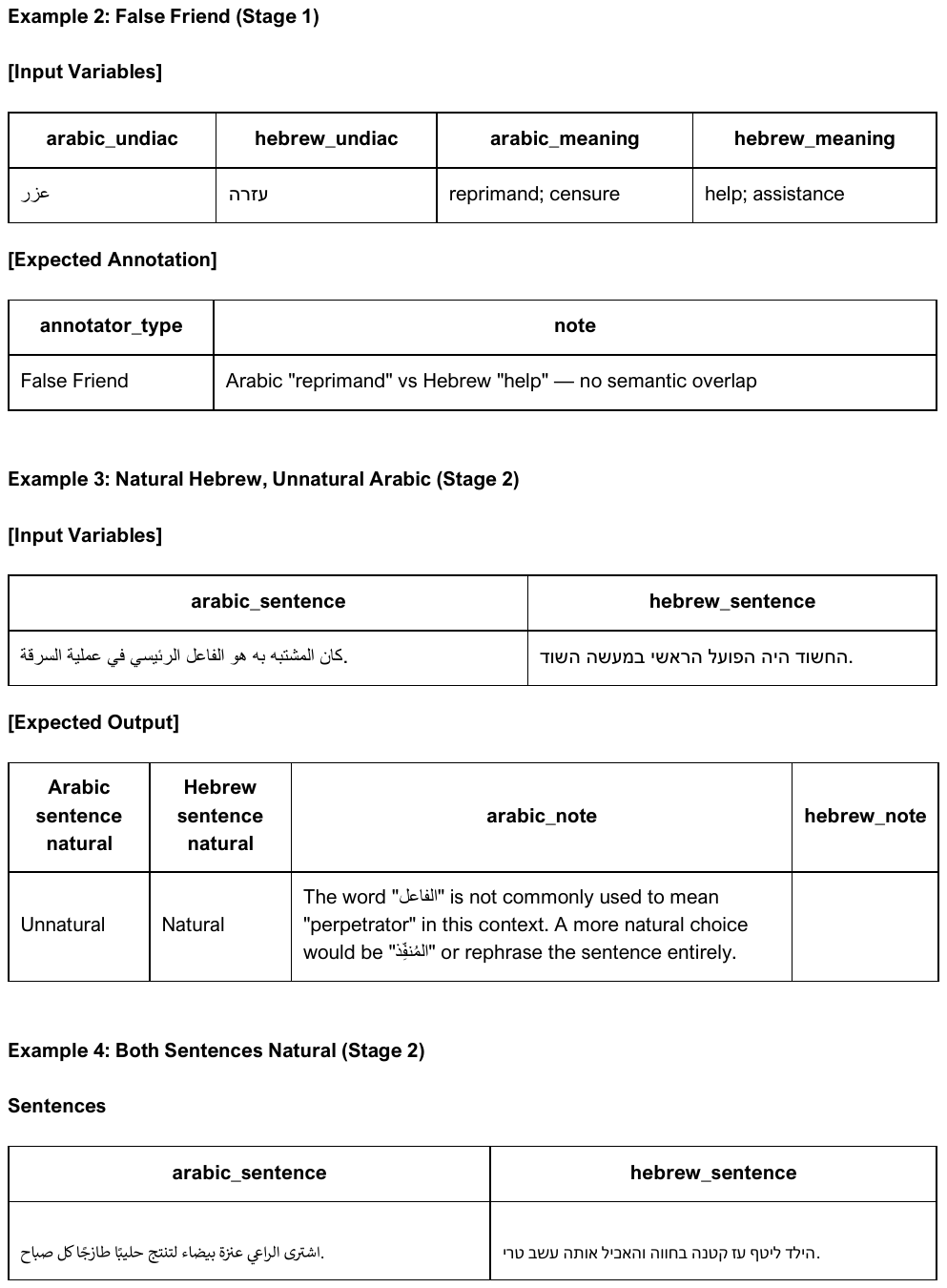}}

    \fbox{\includegraphics[width=\textwidth]{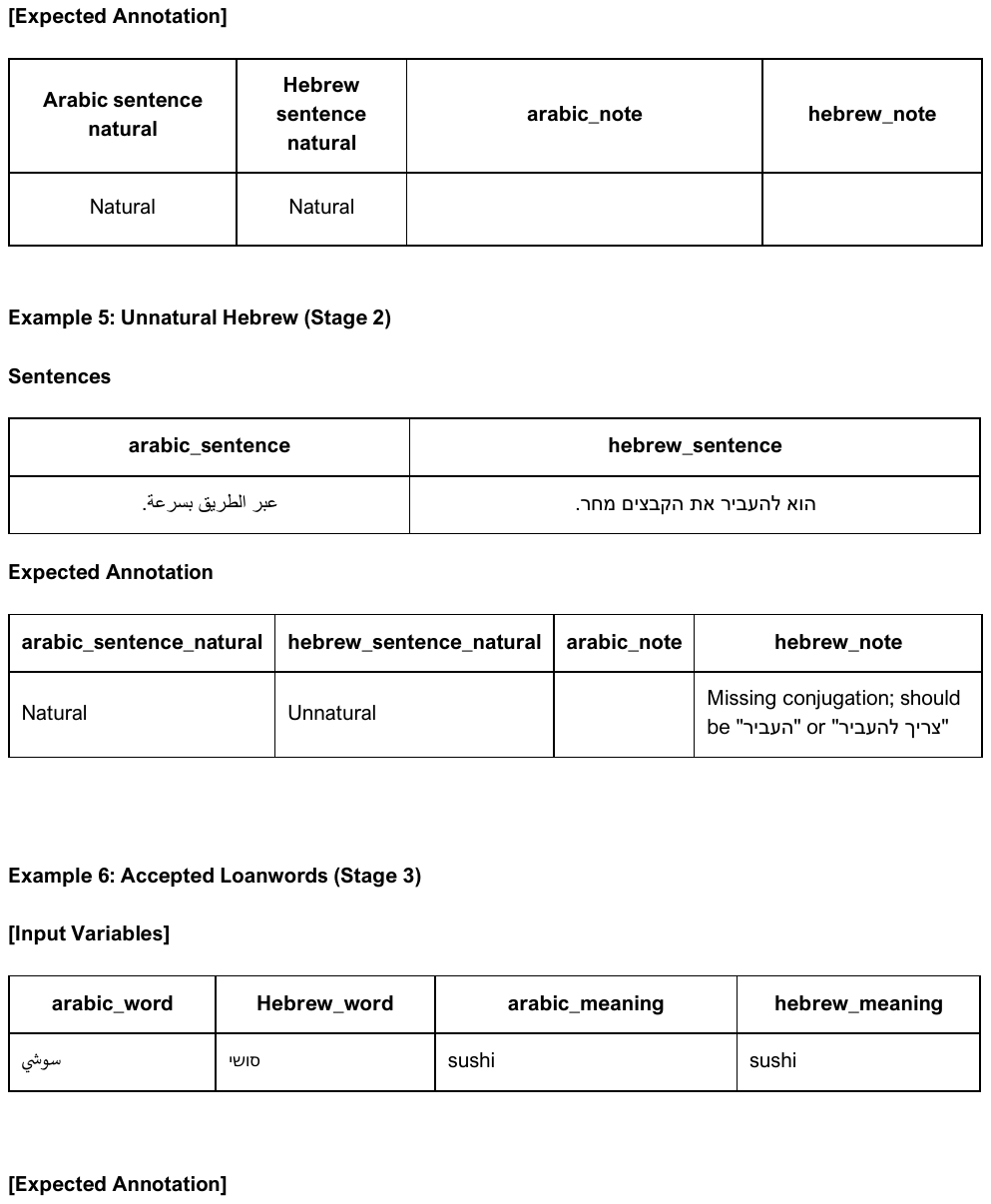}}

    \fbox{\includegraphics[width=\textwidth]{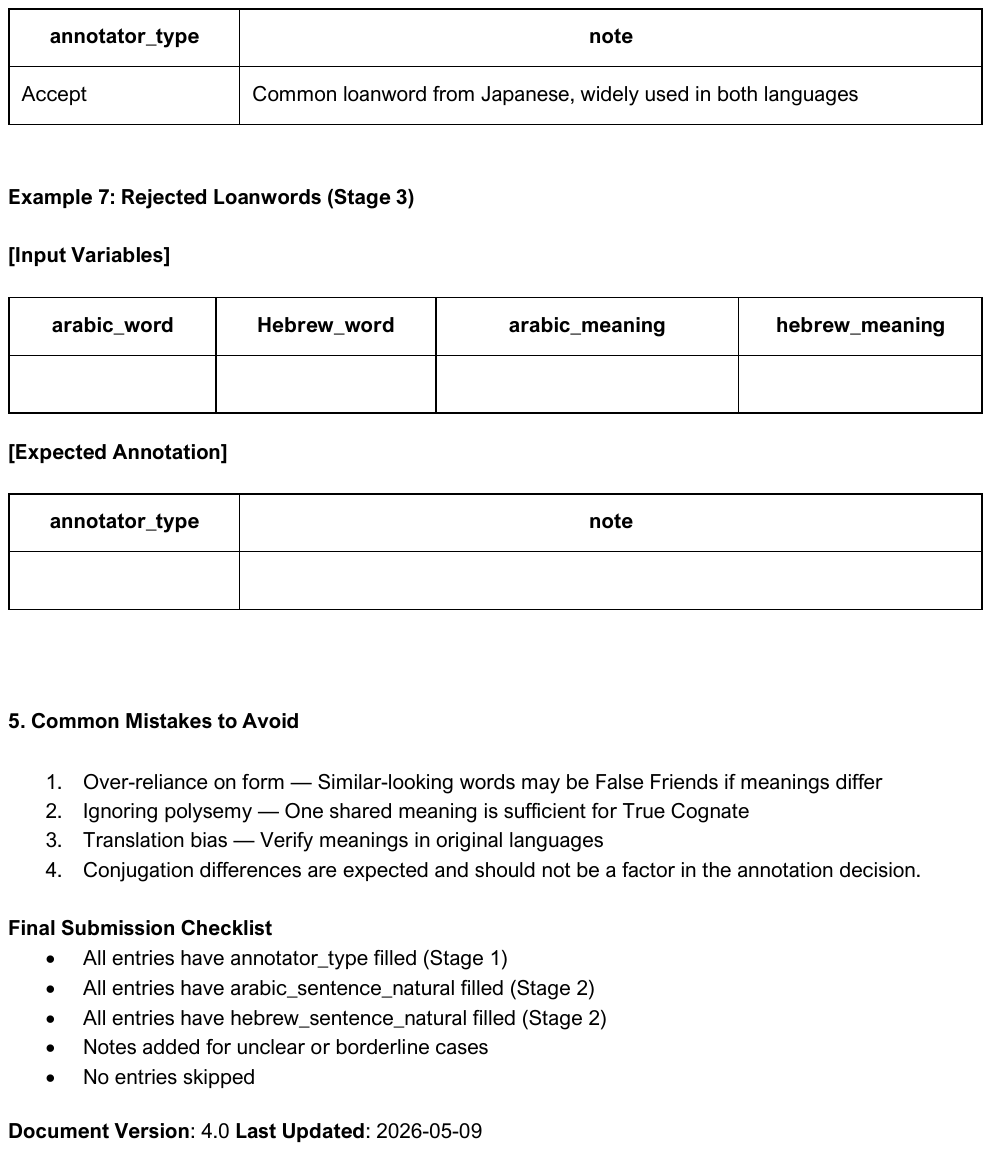}}
\end{center}

\end{document}